\documentclass[runningheads]{llncs}
\usepackage[pagenumbers,year=2024,ID=4105]{eccv}
\usepackage{eccvabbrv}
\usepackage[accsupp]{axessibility}  % Improves PDF readability for those with disabilities.

\usepackage{times}
\usepackage{epsfig}
\usepackage{graphicx}
\usepackage{amsmath}
\usepackage{amssymb}

\usepackage{booktabs}
\usepackage{balance}
\usepackage{multirow}
\usepackage{multicol}
\usepackage{colortbl}
\definecolor{mygray}{gray}{.85}
\definecolor{myhighlight}{RGB}{193,210,240}
\usepackage{mwe}
\usepackage{algorithmic,float}
\usepackage[linesnumbered,ruled,vlined,noline]{algorithm2e}
\usepackage{etoolbox}
\usepackage{relsize}
\usepackage{comment}
\usepackage{bm}
\usepackage{enumitem}
\usepackage{adjustbox}
\usepackage{soul}
\usepackage[dvipsnames]{xcolor}

\definecolor{eccvblue}{rgb}{0.21,0.49,0.74}
\usepackage[pagebackref,breaklinks,colorlinks,citecolor=eccvblue]{hyperref}

\begin{document}

%%%%%%%%% TITLE - PLEASE UPDATE
\title{3D Small Object Detection with Dynamic Spatial Pruning}

\author{Xiuwei Xu\textsuperscript{1}\thanks{Equal contribution. ~~\textsuperscript{\dag}~Corresponding author.}, ~Zhihao Sun\textsuperscript{2*}, ~Ziwei Wang\textsuperscript{3}, ~Hongmin Liu\textsuperscript{2\dag}, ~Jie Zhou\textsuperscript{1}, ~Jiwen Lu\textsuperscript{1\dag}\\
\textsuperscript{1}Tsinghua University \\
~\textsuperscript{2}University of Science and Technology Beijing \\
~\textsuperscript{3}Carnegie Mellon University \\
{\tt\small xxw21@mails.tsinghua.edu.cn;}
{\tt\small d202210361@xs.ustb.edu.cn;} \\
{\tt\small ziweiwa2@andrew.cmu.edu;}
{\tt\small hmliu@ustb.edu.cn;} \\
{\tt\small \{jzhou, lujiwen\}@tsinghua.edu.cn} \\
}

\maketitle
% \twocolumn[{
% \maketitle
% \vspace{-6mm}
% \begin{figure}[H]
% \hsize=\textwidth
% \centering
% \includegraphics[width=2.0\linewidth]{figures/teaser2.pdf}
% \caption{Trained with only rooms from ScanNet, our DSPDet3D generalizes well to process a whole house with dozens of rooms. It takes less than 2s to generate fine-grained detection results with a RTX 3090 single GPU.}
% \label{teaser}
% \end{figure}
% }]

%%%%%%%%% ABSTRACT
\begin{abstract}
  In this paper, we propose an efficient feature pruning strategy for 3D small object detection.
Conventional 3D object detection methods struggle on small objects due to the weak geometric information from a small number of points.
Although increasing the spatial resolution of feature representations can improve the detection performance on small objects, the additional computational overhead is unaffordable.
With in-depth study, we observe the growth of computation mainly comes from the upsampling operation in the decoder of 3D detector.
Motivated by this, we present a multi-level 3D detector named DSPDet3D which benefits from high spatial resolution to achieves high accuracy on small object detection, while reducing redundant computation by only focusing on small object areas. 
Specifically, we theoretically derive a dynamic spatial pruning (DSP) strategy to prune the redundant spatial representation of 3D scene in a cascade manner according to the distribution of objects. Then we design DSP module following this strategy and construct DSPDet3D with this efficient module.
%%%
On ScanNet and TO-SCENE dataset, our method achieves leading performance on small object detection.
Moreover, DSPDet3D trained with only ScanNet rooms can generalize well to scenes in larger scale. It takes less than 2s to directly process a whole building consisting of more than 4500k points while detecting out almost all objects, ranging from cups to beds, on a single RTX 3090 GPU. 
\href{https://xuxw98.github.io/DSPDet3D/}{Code}.
    \vspace{-2mm}
    \vspace{-2mm}

\begin{figure}[t]
    \centering
    \includegraphics[width=1.0\linewidth]{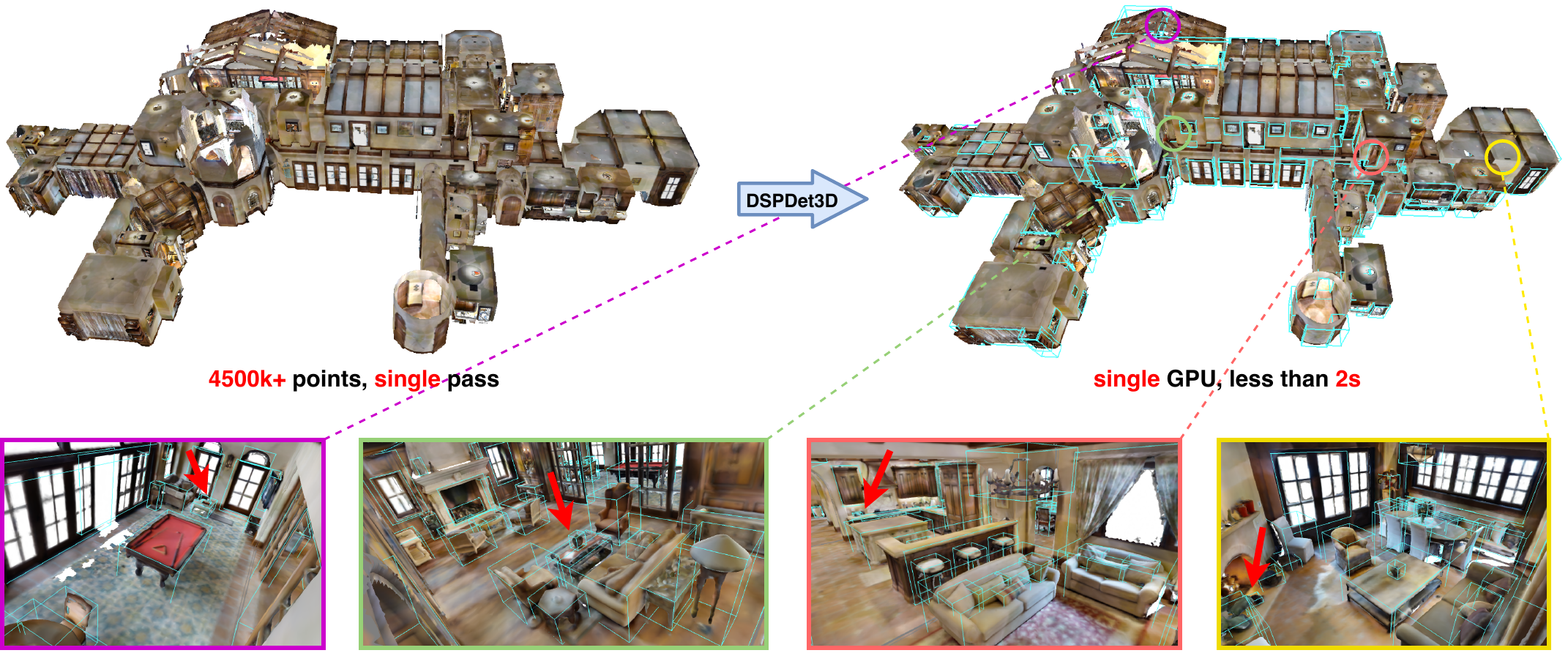}
    \caption{Trained with only rooms from ScanNet, our DSPDet3D generalizes well to process a whole house with dozens of rooms. It takes less than 2s to generate fine-grained detection results with a RTX 3090 single GPU.}
\end{figure}
\end{abstract}

%%%%%%%%% BODY TEXT
\section{Introduction}
  3D object detection is a fundamental scene understanding problem, which aims to detect 3D bounding boxes and semantic labels from a point cloud of 3D scene. With the recent advances of deep learning techniques on point cloud understanding~\cite{qi2017pointnet,qi2017pointnet++,graham20183d,choy20194d}, 3D detection methods have shown remarkable progress~\cite{shi2020pv,zheng2021se,wang2022cagroup3d,rukhovich2023tr3d}. 
However, with 3D object detection being widely adopted in fields like robotics~\cite{zhu2017target,Mousavian_2019_ICCV} and autonomous driving~\cite{bansal2018chauffeurnet} which require highly precise and fine-grained perception, small object detection becomes one of the most important yet unsolved problems.
In autonomous driving scenarios~\cite{geiger2012we}, we observe a significant performance gap between cars and pedestrians. In indoor scenes~\cite{dai2017scannet,Matterport3D} where the size variance is much larger (e.g. a bed is 1000x larger than a cup), detecting small objects is more difficult. 
We focus on indoor 3D object detection task where scenes are crowded with objects of multiple categories and sizes.

For indoor 3D object detection, although great improvement has been achieved in both speed and accuracy on previous benchmarks~\cite{dai2017scannet,armeni20163d,song2015sun}, they are still far from general purpose 3D object detection due to the limited range of object size they can handle. 
For instance, these methods focus on furniture-level objects such as bed and table, while smaller ones like laptop, keyboard and bottle are ignored.
With the arrival of 3D small object benchmarks~\cite{xu2022back,xu2022scene,rozenberszki2022language} which contain objects with wider size variance (e.g.\ from tabletop object like cup to large furniture like bed), it is shown that previous 3D detectors get very low accuracy on small objects and some even fail to detect any small objects. 
This is because extracting fine-grained representation for a large scene is too computationally expensive, so current methods aggressively downsamples the 3D features, which harms the representation of small objects. 

\begin{figure}[t]
    \centering
    \includegraphics[width=0.8\linewidth]{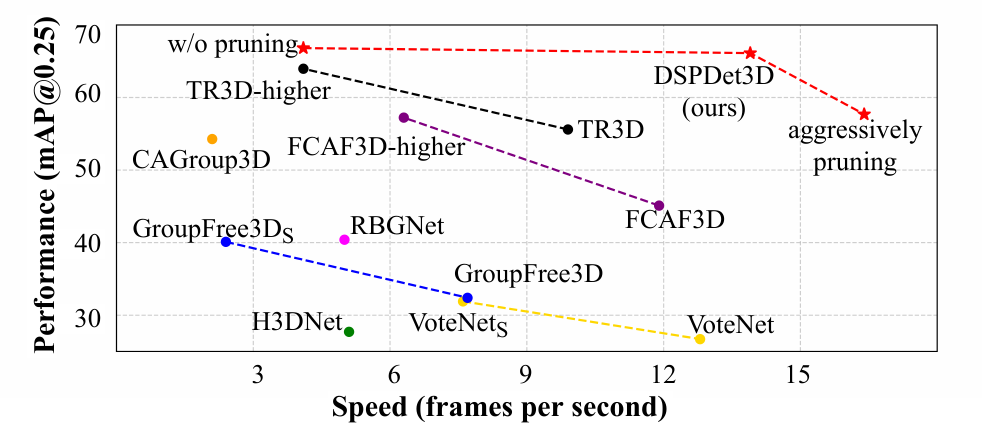}
    \caption{Detection accuracy (mAP@0.25 of all categories) and speed (FPS) of mainstream 3D object detection methods on TO-SCENE dataset. Our DSPDet3D shows absolute advantage on 3D small object detection and provides flexible accuracy-speed tradeoff by simply adjusting the pruning threshold without retraining.}
    \label{teaser2}
\end{figure}

In this paper, we propose a dynamic spatial pruning approach for 3D small object detection. 
Although increasing the spatial resolution of the feature representations is a simple and effective way to boost the performance of 3D small object detection, the large computational overhead makes this plan infeasible for real
application. With in-depth study, we observe the memory footprint mainly comes from the huge number of features generated by the upsampling operation in the decoder of 3D detector.
Inspired by the fact that small objects only occupy a small proportion of space, we adopt a multi-level detection framework to detect different sizes of objects in different levels. 
As the multi-level detector has already detected out larger objects in lower resolution, there are many redundant features in the scene representations of higher resolution.
To this end, we propose to dynamically prune the features after detecting out objects in each level, which skips the upsampling operation at regions where there is no smaller object. 
Specifically, we first theoretically derive a pruning mask generation strategy to supervise the pruning module, which prunes as much features as possible while not affecting the features of object proposals. Then we design a dynamic spatial pruning (DSP) module according to the theoretical analysis and use it to construct a 3D object detector named DSPDet3D.
On the popular ScanNet~\cite{dai2017scannet} dataset, DSPDet3D improves the mAP of all categories by 3\% and mAP of small object by 14\% compared with current state-of-the-art. On TO-SCENE~\cite{xu2022scene} dataset with more tabletop objects, we improve the mAP of all categories by 8\% while achieving leading inference speed among all mainstream indoor 3D object detection methods. 
% We also transfer DSPDet3D trained with only ScanNet rooms to building-level scenes in Matterport3D~\cite{Matterport3D} to demonstrate the generalization ability and efficiency of the proposed approach.
  
\section{Related Work}
  \textbf{Indoor 3D object detection:}
Since PointNet and PointNet++~\cite{qi2017pointnet,qi2017pointnet++}, deep learning-based 3D detection methods for point clouds begin to emerge in recent years, which can be mainly divided into three categories: voting-based~\cite{Qi_2019_ICCV,xie2020mlcvnet,zhang2020h3dnet,cheng2021back,wang2022rbgnet}, transformer-based~\cite{misra2021end,liu2021group} and voxel-based~\cite{gwak2020generative,rukhovich2022fcaf3d,wang2022cagroup3d,rukhovich2023tr3d} methods.
Inspired by 2D hough voting, VoteNet~\cite{Qi_2019_ICCV} proposes the first voting-based 3D detector, which aggregates the point features on surfaces into object center by 3D voting and predicts bounding boxes from the voted centers. Drawing on the success of transformer-based detector~\cite{carion2020end} in 2D domain, GroupFree3D~\cite{liu2021group} and 3DETR~\cite{misra2021end} adopts transformer architecture to decode the object proposals into 3D boxes. As extracting point features require time-consuming sampling and aggregation operation, GSDN~\cite{gwak2020generative} proposes a fully convolutional detection network based on sparse convolution~\cite{graham20183d,choy20194d,lee2021putting,xu2023binarizing}, which achieves much faster speed. FCAF3D~\cite{rukhovich2022fcaf3d} and TR3D~\cite{rukhovich2023tr3d} further improves the performance of GSDN with a simple anchor-free architecture.
Our method also adopts voxel-based architecture considering its efficiency and scalability.

\textbf{Small object detection:}
Small object detection~\cite{tong2020recent} is a challenging problem in 2D vision due to the low-resolution features. To tackle this, a series of methods have been proposed, which can be categorized into three types: (1) small object augmentation and oversampling methods~\cite{kisantal2019augmentation,liu2016ssd,zoph2020learning}; (2) scale-aware training and inference strategy~\cite{singh2018analysis,singh2018sniper,gao2018dynamic,najibi2019autofocus}; (3) increasing the resolution of features or generating high-resolution features~\cite{li2017perceptual,lin2017fpn,chen2017r,wang2020deep,deng2021extended,yang2022querydet}.
However, there are far less works about 3D small object detection due to the limit of data and network capability. BackToReality~\cite{xu2022back} proposes ScanNet-md40 benchmark which contains small objects and finds many current methods suffer a lot in small object detection. TO-SCENE~\cite{xu2022scene} proposes a new dataset and learning strategy for understanding 3D tabletop scenes. However, it relies on densely sampled points from CAD models, which is infeasible in practical scenarios where the points from small objects are very sparse. In contrast, we aim to directly detect small objects from naturally sampled point clouds.

\textbf{Network pruning:}
Network pruning can be divided into two categories: architecture pruning~\cite{lecun1989optimal,han2015learning,molchanov2016pruning,huang2018data,li2016pruning,liu2017learning} and spatial pruning~\cite{rao2021dynamicvit,liu2022spatial,zhao2023ada3d}. Architecture pruning aims to remove a portion of weights from a neural network to shrink the size of a network, which includes unstructured pruning~\cite{lecun1989optimal,han2015learning,molchanov2016pruning} and structured pruning~\cite{huang2018data,li2016pruning,liu2017learning}. The former removes network weights without a predefined structure, while the latter removes whole channels or network layers. On the contrary, spatial pruning does not prune the parameters of a network, but spatially removing redundant computation on the feature maps. DynamicViT~\cite{rao2021dynamicvit} prunes the tokens in vision transformer with an attention masking strategy. SPS-Conv~\cite{liu2022spatial} dynamically prunes the convolutional kernel to supress the activation on background voxels in sparse convolution layer. Ada3D~\cite{zhao2023ada3d} proposes a pruning framework for 3D and BEV features. Our dynamic spatial pruning method also belongs to spatial pruning, which directly removes redundant voxel features level by level according to the distribution of objects.

\section{Approach}
  % sec 4.2 more description
% Fig7/8/9 reorganize
% supplementary material

In this section, we describe our DSPDet3D for efficient 3D small object detection. We first revisit the multi-level 3D detector and analyze the computational cost distribution. Then we propose dynamic spatial pruning with theoretical analysis on how to prune features without affecting detection performance. Finally we design DSP module according to the theoretical analysis and use it to construct DSPDet3D.

\begin{figure}[t]
  \centering
  \includegraphics[width=0.8\linewidth]{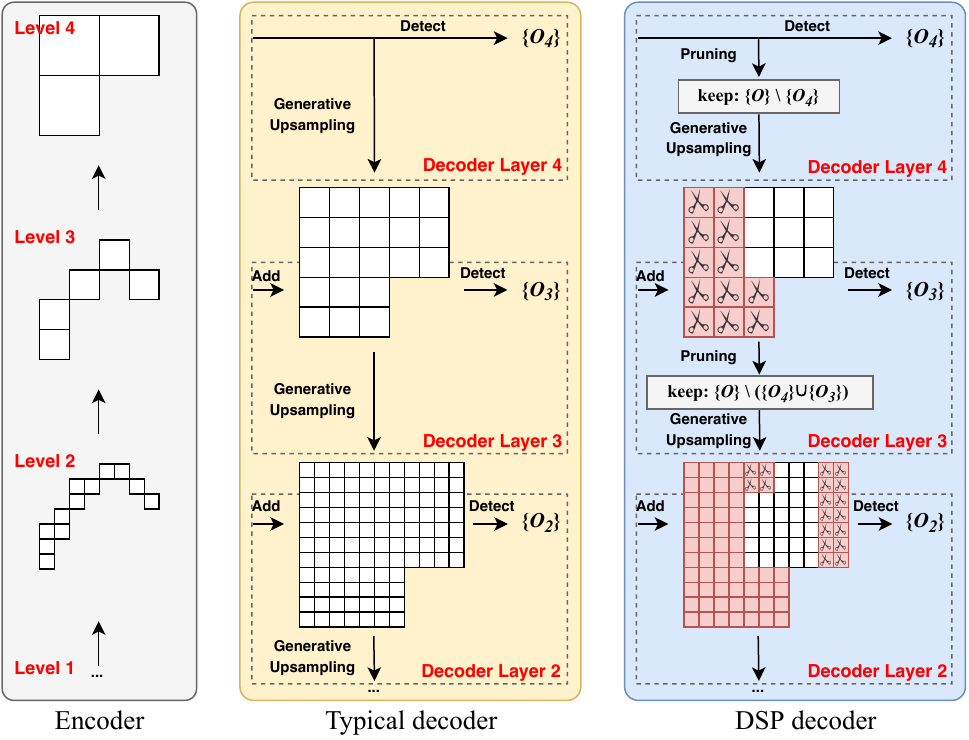}
  \caption{Comparison of the decoder in typical multi-level 3D object detector~\cite{rukhovich2023tr3d} and our DSPDet3D. 
  Note that the sparsity of voxels in decoder is changed due to the generative upsampling operation.
  After detecting out objects in a level, DSPDet3D prunes redundant voxel features according to the distribution of objects before each upsampling operation. Red boxes indicate all pruned voxels and `scissor' boxes indicate voxels pruned in the previous layer. $\{O\}$ is the set of all objects and $\{O_i\}$ is the set of objects assigned to level $i$.}
  \label{motivation}
\end{figure}

\subsection{Analysis on Multi-level 3D Detector}
\textbf{Preliminaries:} We choose multi-level FCOS-like~\cite{tian2019fcos} 3D detector~\cite{rukhovich2022fcaf3d,rukhovich2023tr3d} with sparse convolution~\cite{graham20183d,choy20194d} for small object detection due to its high performance on both accuracy and speed (more detail can be found in Table \ref{tbl:scannet} and \ref{tbl:toscene}).

As shown in Figure \ref{motivation} (middle), after extracting backbone features, multi-level detector iteratively upsamples the voxel feature representations to different levels. In each level, all voxels are regarded as object proposals to predict bounding boxes and category scores. Generative upsampling is widely adopted in this kind of architectures~\cite{gwak2020generative,rukhovich2022fcaf3d,rukhovich2023tr3d} to expand the voxels from object surfaces to the whole 3D space, where object proposals located at object centers can produce accurate predictions.
During training, ground-truth bounding boxes are assigned to different levels and each box assigns several nearby voxels as positive object proposals. 
Only box predictions from positive object proposals will be supervised. While at inference time all voxel features from the decoder are used to predict bounding boxes, which are then filtered by 3D NMS.

\begin{figure}[t]
    \centering
    \includegraphics[width=0.8\linewidth]{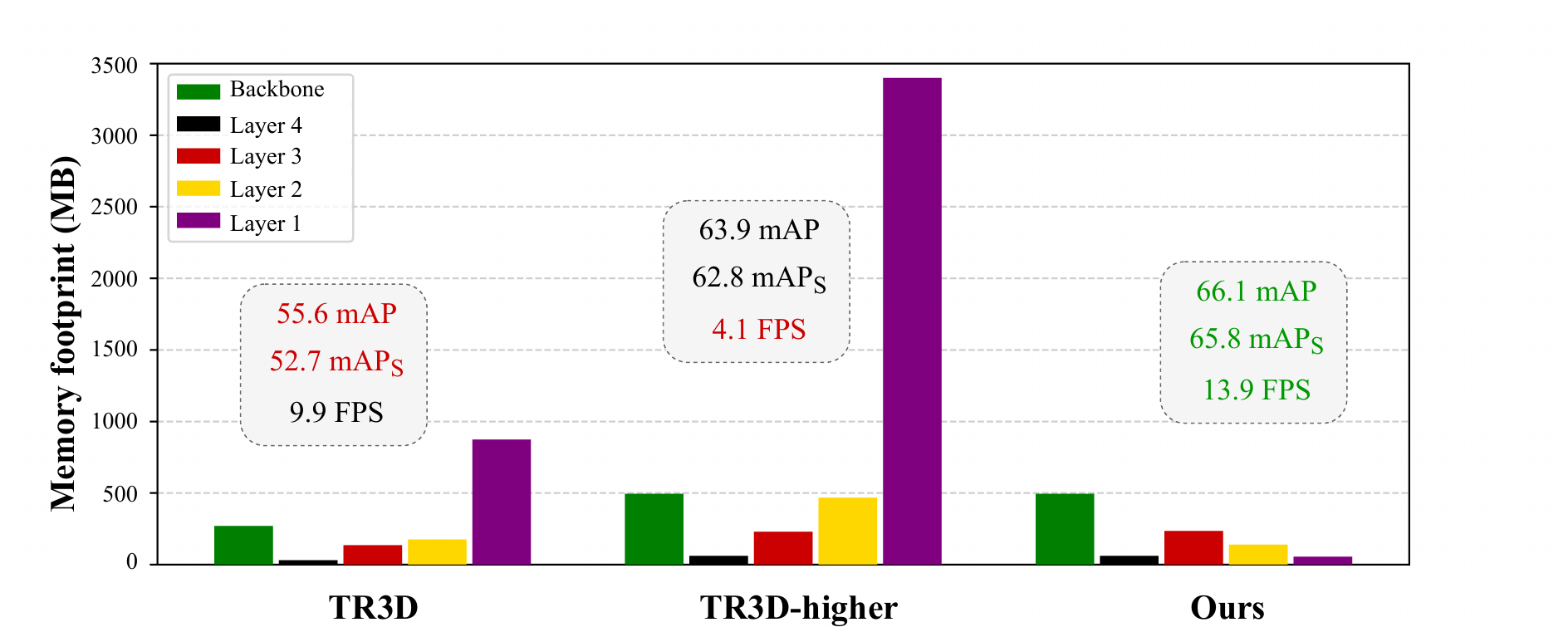}
    \caption{The memory footprint distribution of different multi-level detectors. Layer 4 to Layer 1 refer to decoder layers (including detection heads) from coarse to fine. If doubling the spatial resolution of TR3D, the performance on 3D small object detection improves from 52.7\% to 62.8\% while memory footprint increases dramatically. We find decoder layers accounts for most of the costs. DSPDet3D efficiently reduces redundant computation on these layers, achieving both fast speed and high accuracy.}
    \label{study}
\end{figure}

\textbf{Increasing spatial resolution:} Based on multi-level architecture, a simple way to boost the performance of small object detection is to increase the spatial resolution of feature maps, i.e., voxelizing the point clouds into smaller voxels to better preserve geometric information. Taking TR3D~\cite{rukhovich2023tr3d} for example, we double its spatial resolution and show the results in Figure \ref{study}. It can be seen that the performance on small object really benefits from larger resolution, but the computational overhead grows dramatically at the same time.
As 3D object detection is usually adopted in tasks which requires realtime inference under limited resources, such as AR/VR and robotic navigation, directly increasing spatial resolution is infeasible.
% Infeasible !!!
Notably, we find the computation growth is imbalanced: the decoder layers (including detection heads) account for the most memory footprint and have larger memory growth ratio than the backbone. This indicates the generative upsampling operation will significantly increase the number of voxels when the spatial resolution is high, which is the main challenge for scaling up the spatial resolution of multi-level detectors. 

\subsection{Dynamic Spatial Pruning}\label{theory}
% Idea: small objects only occupy a small proportion of space, while current methods utlize a complete and high-resolution feature map to detect small object
Since small objects only occupy a small proportion of space, we assume there is a large amount of redundant computation in decoder layers, especially when the resolution is high. For instance, if a bed is detected in Layer 4, the region near this bed may be less informative for detecting other objects in the follow decoder layers. If we can skip the upsampling operation at these regions, the voxels will be sparsified level by level, as shown in Figure \ref{motivation} (right). In this way, small objects can be detected in Layer 1 from only a small number of voxels. Inspired by this, we propose to dynamically prune the voxel features according to the distribution of objects.

However, pruning a voxel will not only reduce the number of object proposals in the following levels, but also change the following voxel features computed based on the pruned voxel. Therefore, in order to reduce the redundant computation of multi-level detector without degrading the detection performance, a carefully designed pruning strategy is required. We give theoretical derivation as below.

% more formal, introduct A() as the affecting field
\textbf{Problem formulation:} For each scene, we denote $\{O\}$ as the set of all objects, $\{O_i\}$ as the set of objects assigned to level $i$\footnote{We adopt the same definition of level as in Figure \ref{motivation}, where level $i$ is finer than level $i+1$.} during training, $f_i \in \mathbb{R}^{N\times (3+C)}$ as the voxel features of level $i$. We aim to prune $f_i$ after detecting out $\{O_i\}$, where the objective is to remove as many voxels as possible while keeping the predictions of $\{O\} \backslash \{O_i\}$ unaffected after the pruning.
For each object $o_j$ in level $j\ (j<i)$, we assume the prediction of it is unaffected if the voxel features at level $j$ near its center $\boldsymbol{c_j}$ are unaffected. We make this assumption because most true positive predictions are from object proposals located at the center of bounding boxes~\cite{tian2019fcos,gwak2020generative}. We denote the expected unaffected neighborhood as $\mathcal{C}_j(\boldsymbol{c_j}, P)$, which means a cube centered at $\boldsymbol{c_j}$ with $P\times P\times P$ voxels at level $j$. Given the symmetry, P should be odd.
Then we formulate the objective of our pruning strategy at level $i$ as:
\begin{gather}
  \mathop{\rm minimize}\limits_{\mathcal{K}_i} \sum_{x,y,z}{M_i[x][y][z]}, M_i=\textstyle\bigwedge_{j=1}^{i-1}{\mathcal{K}_i(\boldsymbol{c_j})}, \nonumber \\
  s.t.\ \forall j<i,\ \mathcal{C}_j(\boldsymbol{c_j}, P) \cap \mathcal{A}_{i,j}(\neg\mathcal{K}_i(\boldsymbol{c_j}) \star f_i)=\varnothing \label{eq1}
\end{gather}
where $M_i\in \mathbb{R}^{N}$ is a binary pruning mask sharing the same length with $f_i$, where 0 indicates removing and 1 indicates keeping during the pruning operation $\star$. $\mathcal{K}_i(\cdot)$ is the generation strategy of pruning mask for each object, which generates a binary pruning mask conditioned on the object center. $\mathcal{A}_{i,j}(f)$ is defined as the \emph{affecting field} of $f$, which represents the voxels at level $j$ that will be affected by pruning $f$ at level $i$. Without loss of generality, here we choose only one object at each level for simplicity of presentation.

\textbf{Overview of problem solving:} 
% As solving (\ref{eq1}) is NP-hard and different coordinate sets of $f_i$ results in different $keep(\cdot)$ strategies, we aim to find a strong sufficient condition for $\bigwedge_{j=1}^{i-1} NA(o_j|prune(f_i,keep(c_j)))=True$ which can derive a unified $keep(\cdot)$ strategy.
We solve (\ref{eq1}) by mathematical induction. Specifically, for pruning strategy $M_i$ at level $i$, we first consider how to generate pruning mask $\mathcal{K}_i(\boldsymbol{c_{i-1}})$ to ensure the predictions of $\{O_{i-1}\}$ are unaffected. Then we show that by following our pruning strategy $\mathcal{K}_i$, `the predictions of $\{O_j\}$ are unaffected' can be derived by `the predictions of $\{O_{j+1}\}$ are unaffected'.\footnote{We provide illustrated examples in supplementary material for better understanding.}

\textbf{Solving $\mathcal{K}_i(\boldsymbol{c_{i-1}})$:} To make sure $\mathcal{C}_{i-1}(\boldsymbol{c_{i-1}}, P) \cap \mathcal{A}_{i,i-1}(\cdot)=\varnothing$, we need to compute the affecting field of each voxel $v_i$ in level $i$. Obviously, the upper bound of affecting field of $v_i$ expands in shape of cube with sparse convolution. Assume there are $m$ sparse convolution with stride 1 and kernel $x_k\ (1\leq k\leq m)$ between pruning and generative upsampling in level $i$, one generative transposed convolution with stride 2 and kernel $y$, and $n$ sparse convolution with stride 1 and kernel $z_k\ (1\leq k\leq n)$ until detecting out objects in level $i-1$. Then the affecting field from pruning (level $i$) to detecting (level $i-1$) can be written as:
\begin{gather}
  \mathcal{A}_{i,i-1}(v_i)=\mathcal{C}_{i-1}(v_i, aff(\{x_k\},y,\{z_k\}))
\end{gather}
where $aff(\{x_k\},y,\{z_k\})$ is the range of affecting field represented by the kernel sizes, which we will detail in supplementary material.
% \begin{gather}
%   \mathcal{A}_{i,i-1}(v_i)=\mathcal{C}_{i-1}(v_i, aff(\{x_k\},y,\{z_k\})) \nonumber \\
%   aff(\{x_k\},y,\{z_k\})=\textstyle2\sum_{k=1}^m{x_k}+y+\sum_{k=1}^n{z_k}-2m-n+2
% \end{gather}
Since the shape of the expected unaffected voxel features is a $P\times P\times P$ cube, $\mathcal{K}_i(\boldsymbol{c_{i-1}})$ can be formulated as:
\begin{gather}
  \mathcal{K}_i(\boldsymbol{c_{i-1}})[x][y][z]=\mathbb{I}(2\cdot |\boldsymbol{x}-\boldsymbol{c_{i-1}}|_\infty\leq rS_i) \nonumber \\
  r=\lceil \frac{P+aff(\{x_k\},y,\{z_k\})-2}{2}\rceil \label{eq2}
\end{gather}
where $S_i$ is the size of voxel in level $i$. $\mathbb{I}(\cdot)$ is the indicative function. $\boldsymbol{x}=(x,y,z)$ is the voxel coordinates of $f_i$.

\textbf{Recursion of $\mathcal{K}_i$:} We now derive when the pruning strategy $\mathcal{K}_i$ in (\ref{eq2}) also works for $\boldsymbol{c_j}\ (j<i-1)$. We can regrad $\boldsymbol{c_j}$ as the center of object in level $i-1$ and use (\ref{eq2}) to generate the pruning mask. In this way, $\mathcal{C}_{i-1}(\boldsymbol{c_j}, P)$ are unaffected. As $\mathcal{C}_{j}(\boldsymbol{c_j}, P)$ is covered by $\mathcal{C}_{i-1}(\boldsymbol{c_j}, P)$, so $\mathcal{C}_{j}(\boldsymbol{c_j}, P)$ is unaffected as well. We should also ensure pruning in level $i$ has no cumulative impact on pruning in level $i-1$:
\begin{equation}
  (\mathcal{K}_{i-1}(\boldsymbol{c_j}) \star f_{i-1}) \subseteq \mathcal{C}_{i-1}(\boldsymbol{c_j}, P)
\end{equation}
this means when generating pruning mask of $\boldsymbol{c_j}$ in level $i-1$ using $\mathcal{K}_{i-1}$, the kept voxels should be covered by the unaffected voxels after pruning in level $i$. So we have:
\begin{equation}\label{eq3}
  r\cdot S_{i-1} \leq P\cdot S_{i-1}
\end{equation}
The minimum $P$ can be acquired by solving (\ref{eq3}). In this case, strategy $\mathcal{K}_{i}$ in (\ref{eq2}) works for all $\boldsymbol{c_j}\ (j<i)$.

\subsection{DSPDet3D}
Based on the theoretical analysis, we devise a dynamic spatial pruning (DSP) module to approximate the ideal pruning strategy.
We further construct a 3D small object detector named DSPDet3D with the proposed DSP module.

\begin{figure*}[t]
  \centering
  \includegraphics[width=1.0\linewidth]{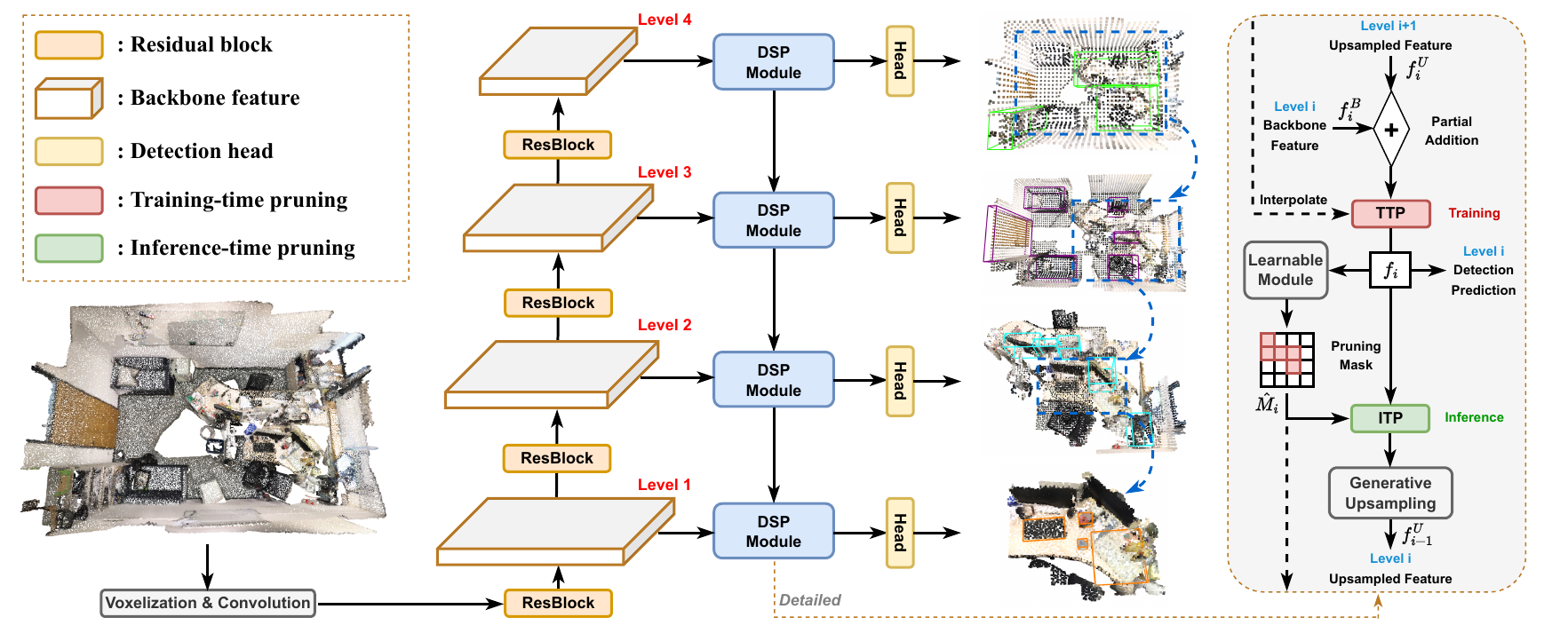}
  \caption{Illustration of DSPDet3D. The voxelized point clouds are fed into a high-resolution sparse convolutional backbone, which output four levels of scene representations. Four dynamic spatial pruning (DSP) modules are stacked to construct a multi-level decoder and detect objects from coarse to fine. DSP module utilizes a light-weight learnable module to predict the pruning mask. During inference, we discretize the pruning mask and use it to guide pruning before generative upsampling. While during training we interpolate the pruning mask to next level and prune the voxel features after generative upsampling.}
  \label{network}
\end{figure*}

\textbf{DSP module:} As shown in Figure \ref{motivation}, we modify the layers of a typical multi-level decoder to DSP modules, which prunes redundant voxel features after detecting out objects at each level for efficient feature upsampling. 
Formally, given the upsampled voxel feature $f^U_i$ and the backbone feature $f^B_i$ at level $i$, DSP module first add them for detection. However, $f^U_i$ may be much sparser than $f^B_i$ due to pruning, directly adding by taking union of them is inefficient. Therefore, we propose a new operator called \emph{partial addition} to fit our pruning strategy:
\begin{equation}
  f_i=f_i^B\overrightarrow{+}f_i^U
\end{equation}
where addition is constrained to be operated only on the voxels of $f_i^U$. Then objects are detected using a shared detection head across all levels: $\{O_i\}=Detect(f_i)$. 
Once objects at level $i$ are detected out, we prune the voxel features according to the derived strategy described in Section \ref{theory}.
Here we devise a light-weight MLP-based learnable pruning module to decide where smaller objects (i.e.\ objects in level $j\ (j<i)$) may appear, and then prune other locations:
\begin{equation}
  \bar{f}_i=t(\hat{M}_i)\star f_i,\ \hat{M}_i={\rm MLP}_i(f_i)
\end{equation}
where $\hat{M}_i$ is the pruning mask predicted from $f_i$, which represents the probability of retention for each voxel. 
We utlize FocalLoss~\cite{lin2017focal} to supervise $\hat{M}_i$ with the generated $M_i$ in (\ref{eq1}).
During inference, a threshold function $t(\cdot)$ sets probability lower than $\tau$ to be 0, others be 1 to guide pruning. 
After pruning, the generative upsampling is applied to acquire features for the next level: $f^U_{i-1}=GeConv(\bar{f}_i)$.

During training, as $\hat{M}_i$ may not be so accurate (especially at beginning), we find applying the above learnable pruning module makes training difficult to converge. Instead, we switch the pruning to weak mode for context preservation.
As shown in Figure \ref{network}, the weak pruning is applied after generative upsampling. For level $i$, we upsample the pruning mask $\hat{M}_{i+1}$ to level $i$ with nearest neighbor interpolation. Then we sort the interpolated scores and keep only $N_{max}$ voxels with the highest scores. This weak pruning mechanism aims to stablize training, which only works when the amount of voxels is too large to conduct following operations.

Since our theoretical analysis sets the expected unaffected neighborhood to be a $P\times P\times P$ cube, we also modify the assigning strategy of positive object proposals accordingly for robust training.
Specifically, for a ground-truth bounding box of $o_i$ assigned to level $i$, we sample the nearest $N_{pos}$ voxels to $\boldsymbol{c_i}$ inside the cube centered at $\boldsymbol{c_i}$ with length $P\cdot S_i$. If there are less than $N_{pos}$ voxels in the cube, we simply sample all voxels inside it.
Our assigning method is independent of the size of bounding box, which ensures there are enough positive proposals even for small objects.

\textbf{DSPDet3D:} Based upon the top-performance multi-level detector TR3D~\cite{rukhovich2023tr3d}, we remove the max pooling layer to increase the spatial resolution of backbone features. Then we replace the decoder in TR3D with four stacked DSP modules to remove redundant voxel features level by level, which achieves efficient upsampling without affecting the detection performance.
To train DSPDet3D, we keep the same loss for classification and box regression as in TR3D and add additional FocalLoss to supervise $\hat{M}_i$ with $M_i$.

\textbf{Compare with FCAF3D:} Similar to our training-time weak pruning, FCAF3D~\cite{rukhovich2022fcaf3d} also adopts a pruning strategy in the decoder to prevent the number of voxels from getting too large, which is unable to remove redundant features in early decoder layers during inference. Moreover, it directly utilizes the classification scores for bounding boxes to sort and prune the voxel features, which cannot accurately preserve geometric information for small objects.
  
\section{Experiment}\label{exp}
  In this section, we conduct experiments to investigate the performance of our approach on 3D small object detection. We first describe the datasets and experimental settings. Then we compare DSPDet3D with the state-of-the-art 3D object detection methods. We also design ablation experiments to study the effectiveness of the proposed methods. Finally we transfer DSPDet3D to extremely large scenes to show its efficiency and generalization ability.

\subsection{Experimental Settings}
\textbf{Datasets and metrics:} We conduct experiments on two indoor datasets including ScanNet~\cite{dai2017scannet} and TO-SCENE~\cite{xu2022scene}. ScanNet is a richly annotated dataset of indoor scenes with 1201 training scenes and 312 validation scenes. Each object in the scenes are annotated with texts and then mapped to category IDs. We follow the ScanNet-md40 benchmark proposed by \cite{xu2022back}, which contains objects in 22 categories with large size variance. TO-SCENE is a mixed reality dataset which provides three variants called TO$\_$Vanilla, TO$\_$Crowd and TO$\_$ScanNet with different numbers of tabletop objects and scene scales. We choose the room-scale TO$\_$ScanNet benchmark, which contains 3600 training scenes and 800 validation scenes with 70 categories. However, TO$\_$ScanNet adopts non-uniform sampling to acquire about 2000 points per tabletop object, which is infeasible in practical settings. To this end, we downsample the small objects and control the density of them to be similar with other objects and backgrounds. We name this modified version as TO-SCENE-down benchmark. We take the point clouds without color as inputs for all methods. More details about ScanNet-md40 and TO-SCENE-down benchmarks can be found in supplementary material.

\begin{table*}[t]
  \setlength{\tabcolsep}{4pt}
  \centering
  \footnotesize
  \caption{
    3D objects detection results and computational costs of different methods on ScanNet-md40. DSPDet3D with the best pruning threshold is highlighted in gray. We set best scores in bold, runner-ups underlined.}
    \begin{tabular}{cccccccc}  
      \toprule
      \multirow{2}{*}{Method} & \multirow{2}{*}{Decoder} & \multicolumn{2}{c}{mAP} & \multicolumn{2}{c}{mAP$_S$} & \multirow{2}{*}{Speed} & \multirow{2}{*}{Memory} \\ 
      & &@0.25 &@0.5 &@0.25 &@0.5 & & \\
      \midrule
      VoteNet & Voting & 51.02 &33.69 & 0.30 &0 & \textbf{13.4} & 1150 \\
      VoteNet$_S$ & Voting & 48.62 &31.55 & 1.04 &0 & 8.5 & 1500 \\
      H3DNet & Hybrid & 53.51 &39.23 & 3.08 &0.90 & 7.2 & 1550 \\
      GroupFree3D & Transformer & 56.77 &41.39 & 11.7 &0.81 & 7.8 & 1450 \\
      GroupFree3D$_S$ & Transformer & 29.44 &11.94 & 0.20 &0 & 3.2 & 2000 \\
      RBGNet & Voting & 55.23 &32.64 & 5.81 &0 & 6.6 & 1700 \\
      FCAF3D & Multi-level & 59.49 &48.75 & 18.38 &8.21 & 12.3 & \underline{850} \\
      CAGroup3D & Voting & 60.29 &49.90 & 16.62 &8.63 & 3.1 & 3250 \\
      TR3D & Multi-level & 61.59 &49.98 & 27.53 &12.91 & 10.8 & 1250 \\
      \midrule
      FCAF3D-higher & Multi-level & 62.65 &51.01 & 27.68 &16.23 & 7.1 & 4000 \\
      TR3D-higher & Multi-level & \underline{65.18} &54.03 & 41.70 &29.56 & 5.2 & 4450 \\
      Ours($\tau=0$) & Multi-level & \textbf{65.39} &\textbf{54.59} & \textbf{44.79} &\textbf{31.55} & 4.4 & 4200 \\
      \rowcolor{mygray} Ours($\tau=0.3$) & Multi-level & 65.04 &\underline{54.35} & \underline{43.77} &\underline{30.38} & \underline{12.5} & \textbf{700} \\
      \bottomrule
    \end{tabular}
  \label{tbl:scannet}
\end{table*}

We report the mean average precision (mAP) with threshold 0.25 and 0.5. To measure the performance on different categories, we use two kinds of metrics: mAP and mAP$_S$, which refer to the mean AP of all objects and of small objects respectively. Here we define categories of small object as ones with average volume smaller than $0.05 m^3$ for both benchmarks.

\textbf{Implementation details:} We implement our approach with PyTorch~\cite{paszke2019pytorch}, Minkow- skiEngine~\cite{choy20194d} and MMDetection3D~\cite{contributors2020mmdetection3d}. We follow the same training strategy / hyperparameters as TR3D~\cite{rukhovich2023tr3d} for fair comparison. Training converges within 4 hours on a 4 GPU machine.
The stride of the sparse convolution in the preencoder of DSPDet3D is set to 2, thus the voxel size of $f_1^B$ is \emph{4cm} and $S_i$ equals to $2^i\cdot$ \emph{2cm}. We set $N_{pos}=6$ and $N_{max}=100000$ during training. The weight of the FocalLoss between $M_i$ and $\hat{M}_i$ is 0.01. In terms of block structure, we have $\{x_k\}=\varnothing$, $y=3$ and $\{z_k\}=\{3,3\}$. So we set $r=7$ and $P=7$ according to (\ref{eq2}).

\begin{table*}[t]
  \setlength{\tabcolsep}{4pt}
  \centering
  \footnotesize
  \caption{
    3D objects detection results and computational costs of different methods on TO-SCENE-down benchmark. DSPDet3D with the best pruning threshold is highlighted in gray. We set best scores in bold, runner-ups underlined.}
    \begin{tabular}{cccccccc}  
      \toprule
      \multirow{2}{*}{Method} & \multirow{2}{*}{Decoder} & \multicolumn{2}{c}{mAP} & \multicolumn{2}{c}{mAP$_S$} & \multirow{2}{*}{Speed} & \multirow{2}{*}{Memory}\\ 
      & &@0.25 &@0.5 &@0.25 &@0.5 & & \\
      \midrule
      VoteNet & Voting & 26.72 &14.01 & 14.51 &4.78 & \underline{12.8}& 1300 \\
      VoteNet$_S$ & Voting & 31.87 &14.89 & 21.75 &7.40 & 7.6& 1650 \\
      H3DNet & Hybrid & 27.69 &17.38 & 14.83 &7.39 &5.1& 1650 \\
      GroupFree3D & Transformer & 32.41 &20.43 & 20.17 &10.13 & 7.7& 1700 \\
      GroupFree3D$_S$ & Transformer & 40.14 &23.55 & 33.33 &16.15 & 2.4& 2200 \\
      RBGNet & Voting & 40.42 &30.27 & 29.69 &21.61 & 5.0& 1850 \\
      FCAF3D & Multi-level & 45.13 &37.21 & 37.18 &31.65 & 11.9& \underline{1000} \\
      CAGroup3D & Voting & 54.28 &47.58 & 48.49 &43.85 & 2.2& 3500 \\
      TR3D & Multi-level & 55.58 &45.95 & 52.72 &44.01 & 9.9& 1400 \\
      \midrule
      FCAF3D-higher & Multi-level & 57.23 &50.39 & 53.07 &48.76 & 6.3& 4250\\
      TR3D-higher & Multi-level & 63.96 &56.06 & 62.84 &57.14 & 4.1& 4600\\
      Ours($\tau=0$) & Multi-level & \textbf{66.81} &\textbf{59.41} & \textbf{66.53} &\textbf{61.57} & 4.1& 5300 \\
      \rowcolor{mygray} Ours($\tau=0.5$) & Multi-level & \underline{66.12} &\underline{58.55} & \underline{65.82} &\underline{60.73} & \textbf{13.9}& \textbf{800} \\
      \bottomrule
    \end{tabular}
  \label{tbl:toscene}
\end{table*}

\subsection{Comparison with State-of-the-art}
We compare our method with popular and state-of-the-art 3D object detection methods, including VoteNet~\cite{Qi_2019_ICCV}, H3DNet~\cite{zhang2020h3dnet}, GroupFree3D~\cite{liu2021group}, RBGNet~\cite{wang2022rbgnet}, CAGroup3D~\cite{wang2022cagroup3d}, FCAF3D~\cite{rukhovich2022fcaf3d} and TR3D~\cite{rukhovich2023tr3d}. We also follow \cite{xu2022scene} to reduce the radius of ball query in the PointNet++ backbone for VoteNet and GroupFree3D. The modified models is distinguished by subscript $S$. 
Note that the original TR3D only uses two detection head at level 2/3 and removes the last generative upsampling. However, detecting small objects heavily relies on high-resolution feature map, so we add the upsampling back. This will make it slightly slower but much more accurate on the 3D small object detection benchmarks.

For all methods, we use their official code and the same training strategy / hyperparameters to train them on ScanNet-md40 and TO-SCENE-down.

\begin{figure*}[t]
  \centering
  % \vspace{-2mm}
  \includegraphics[width=1.0\linewidth]{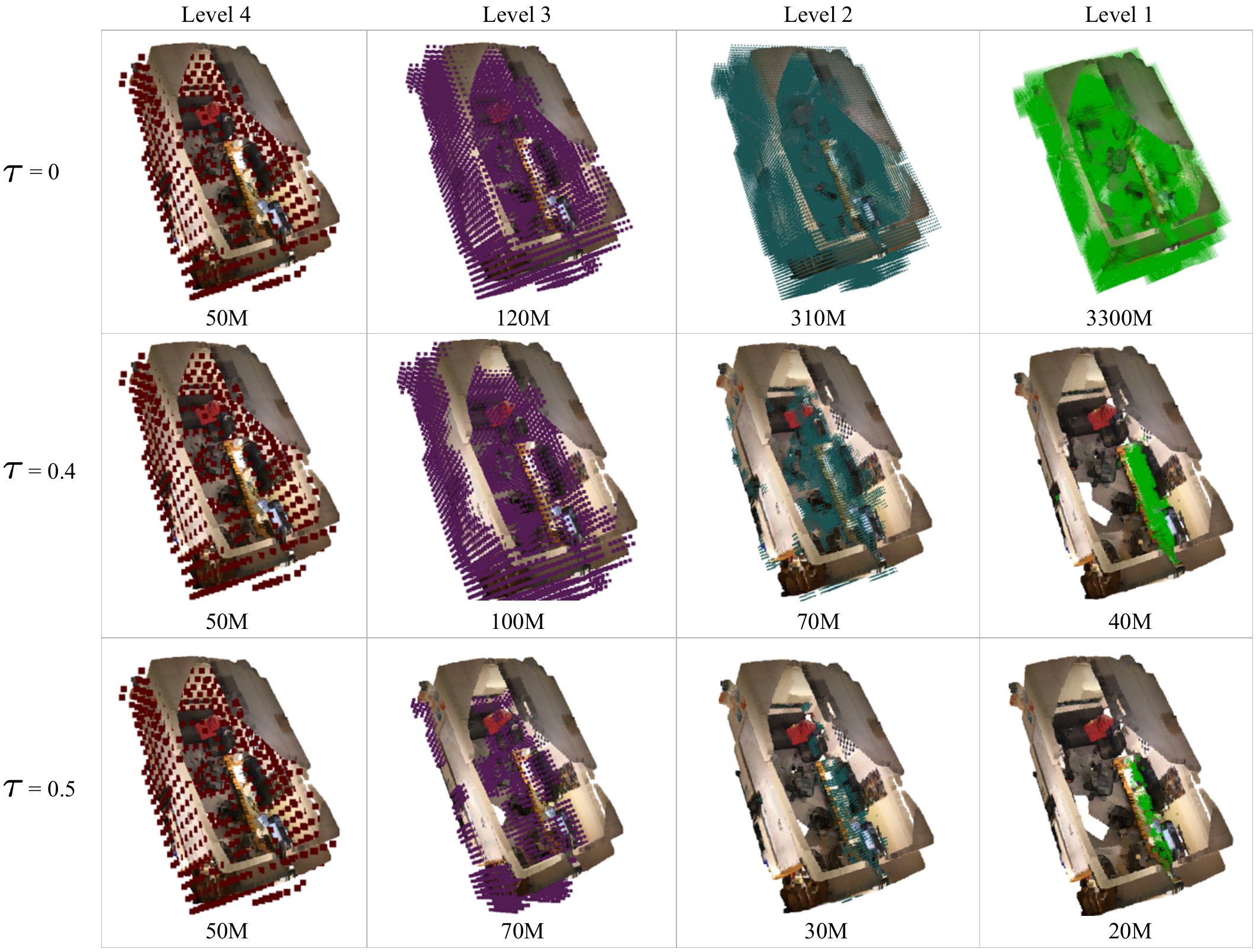}
  \caption{Visualization of pruning process on ScanNet. We show the kept voxels in each level under different thresholds. The memory footprint of each level is also listed at bottom.}
  \label{vis:prune}
\end{figure*}

Table \ref{tbl:scannet} and \ref{tbl:toscene} shows the experimental results on ScanNet-md40 and TO-SCENE-down respectively. 
Consistent with the observation of \cite{xu2022back}, we find point-based (VoteNet, H3DNet, RBGNet) and transformer-based (GroupFree3D) methods almost fail to detect small objects on ScanNet-md40. This is because the PointNet++ backbone used by these methods adopts set abstraction (SA) operation to aggressively downsample the point clouds and extract scene representation. Since the number of small objects in ScanNet is limited, furthest point sampling has a low probability to sample points on small objects, which leads to inaccurate representation of small objects.
For methods (CAGroup3D, FCAF3D, TR3D) with sparse convolutional backbone, they achieve relatively much higher mAP$_S$ due to sparse convolution~\cite{graham20183d,choy20194d} can extract fine-grained scene representation with high efficiency. However, two-stage method like CAGroup3D is both slow and memory-consuming. Multi-level methods like FCAF3D and TR3D are efficient and get good performance on small object detection due to the FPN-like architecture, but they are still limited by resolution. On the contrary, our DSPDet3D with a proper threshold takes advantage of the high-resolution scene representation to achieve much higher performance. Furthermore, DSPDet3D is the most memory-efficient model among all mainstream methods.

\subsection{Ablation Study}
% TODO: this subsection needs to be updated
We conduct ablation studies on ScanNet-md40 to study the effects of hyperparameters and different design choices.

\begin{table}[]
  \centering
  \setlength\tabcolsep{8pt}
  \caption{Ablation studies on several design choices. We control the speed of each method to 10 FPS and report the accuracy in mAP@0.25 and mAP$_S$@0.25.}
  \label{tab:ablation}
  \begin{tabular}{l|p{0.6cm}<{\centering}p{0.6cm}<{\centering}}
      \toprule
      Method & mAP & mAP$_S$ \\
      \midrule
      Remove partial addition &55.3 &35.5 \\
      Addition by taking union &57.9 &36.4 \\
      Addition by interpolation &62.1 &40.9 \\
      Spherical keeping mask &63.0 & 41.1 \\
      Remove training-time pruning &-- &-- \\
      Positive proposal inside bounding box &62.4 &40.7 \\
      \textbf{The full design of DSP module} &\textbf{65.1} &\textbf{44.1} \\
      \bottomrule
  \end{tabular}
\end{table}

\begin{figure}[t]
  \centering
  \includegraphics[width=0.8\linewidth]{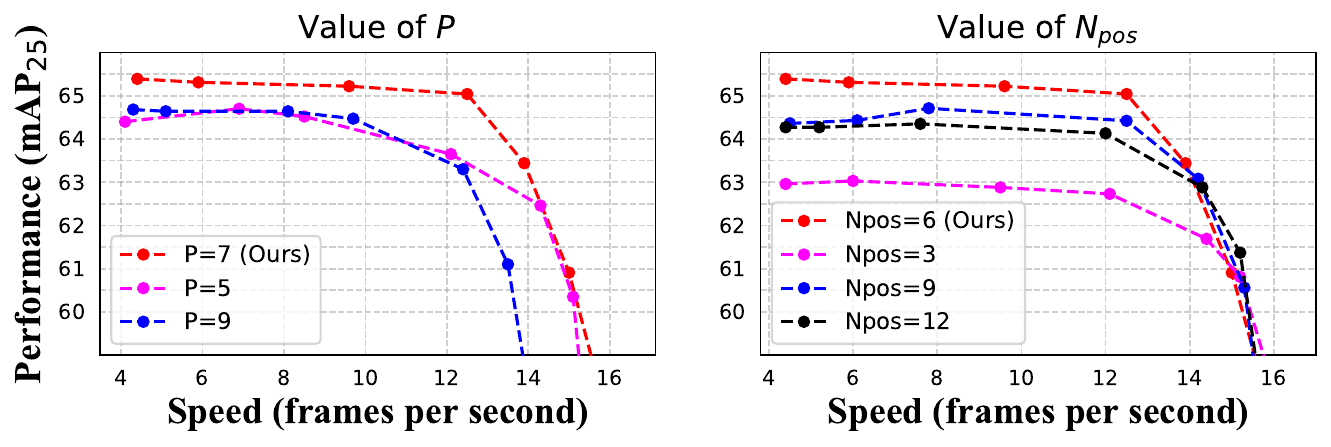}
  \caption{Ablation studies on the value of $r$ and $N_{pos}$. For each value we report performance under different pruning threshold $\tau$.}
  \label{fig:ablation}
\end{figure}

\begin{figure*}[t]
  \centering
  \includegraphics[width=1.0\linewidth]{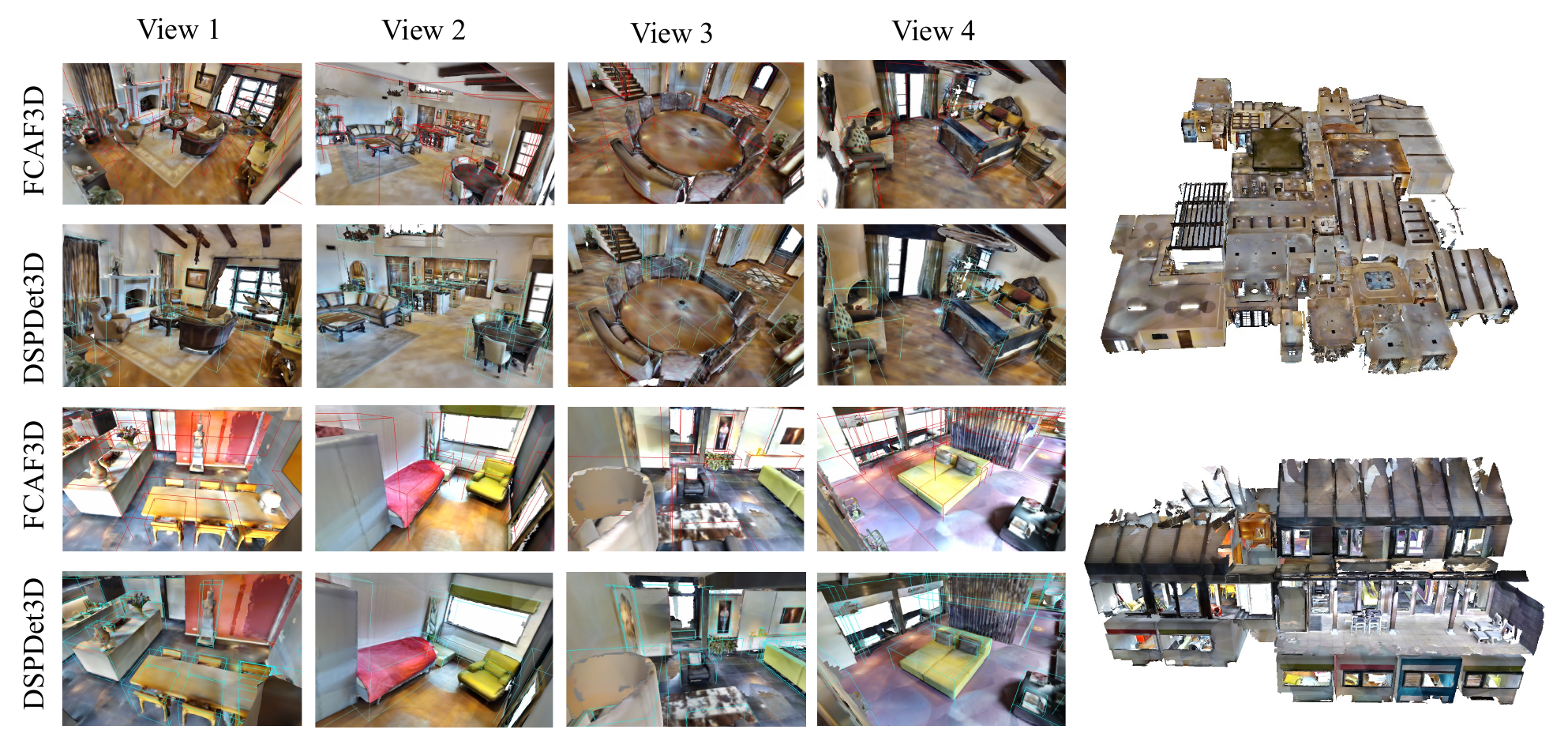}
  \caption{
    Visualization of the transferring results of different 3D object detectors. The 3D detector is trained on rooms from ScanNet and directly adopted to process a whole building-level 3D scene from Matterport3D.}
  \label{vis-house}
\end{figure*}

\textbf{Pruning process:} We visualize the pruning process under different thresholds in Figure \ref{vis:prune}, where the voxels in each level after pruning are shown. We also list the memory footprint of each level. It can be seen that our method significantly reduce the memory footprint by pruning most of the uninformative voxels. Our pruning module only keeps regions where there are smaller objects than current level.

\textbf{Hyperparameters:} We study two hyperparameters: $r$ and $N_{pos}$, which is highly relevant to 3D small object detection. Note that $r=\lceil \frac{P+9-2}{2}\rceil$, thus $r$ and $P$ should be changed simultaneously. As shown in Figure \ref{fig:ablation} (left), setting $r=7$ achieves the best performance. If $r$ is smaller than $7$ then $r>P$, which conflicts with Equation (\ref{eq3}) and the features will be affected by pruning. While a larger $r$ will make the pruning less aggressive, resulting in a large number of redundant voxel features. Figure \ref{fig:ablation} (right) shows that the number of positive object proposals should be set properly, which is important to balance the ratio between positive and negative samples during classification.

\textbf{Design choices:} We also study the design choices of DSPDet3D in Table \ref{tab:ablation}. Observing the second, third and fourth rows, we conclude that the partial addition is important for efficient feature fusion. Although taking union can preserve more information, this operation will reduce the sparsity of voxels and thus make our pruning less efficient. The fifth row shows that generate the keeping mask according to the shape of affecting field is better than using a spherical shape. According to the sixth row, removing training-time pruning will significantly increase the memory footprint during training, which makes the network unable to train. The seventh row validates the effectiveness of our assigning method for positive object proposals.

\subsection{Transferring to Larger Scenes}\label{tls}
We further validate the efficiency and generalization ability of different 3D detectors by transferring them to scenes of much larger scale. We first train 3D detectors on rooms from ScanNet training set in a category-agnostic manner, which is done by regarding every labeled object as the same category. Then we directly adopt them to process the building-level scenes in Matterport3D~\cite{Matterport3D}. 
We find previous methods almost all fail to process the extremely large scenes due to unaffordable memory footprint, so we only compare DSPDet3D with FCAF3D as shown in \ref{vis-house}.
It is shown that FCAF3D cannot detect out any small object and even struggles on relatively large objects like chairs when the scene is too large.
On the contrary, DSPDet3D is able to accurately detect small objects like cups and thin pictures.

% \subsection{Limitation}
% There are two main limitations of DSPDet3D. First, although it achieves high efficiency, the training cost is still larger than inference. We find using the learnable pruning module during training will lead to a large performance drop and so we have to adopt a weak pruning strategy. By solving this problem, DSPDet3D can be trained on scenes of larger scales in less time. Second, currently the input of DSPDet3D is a reconstructed point clouds. We will work on extending it to online 3D detection with RGB-D videos as input, which can support a wider range of practical application.
  
\section{Conclusion}
  In this paper, we have presented an efficient feature pruning strategy for 3D small object detection.
% We manage to boost the performance of 3D detector on small objects by increasing the spatial resolution, while avoiding the unaffordable computational cost brought by a large number of features.
Inspired by the fact that small objects only occupy a small proportion of space, we adopt a multi-level detection framework to detect different sizes of objects in different levels. Then we present a dynamic spatial pruning strategy to prune the voxel features after detecting out objects in each level. 
% In this way, we remove most of the redundant features in high-resolution feature maps and make small object detection much more efficient.
Specifically, we first design the dynamic spatial pruning strategy by theoretical analysis on how to prune voxels without affecting the features of object proposals. Then we propose dynamic spatial pruning (DSP) module according to the strategy and use it to construct DSPDet3D.
Extensive experiments on ScanNet and TO-SCENE datasets show that our DSPDet3D achieves leading detection accuracy and speed. We also conduct transferring experiment on Matterport3D to show DSPDet3D also generalizes well to extremely large scenes.

\section*{Acknowledgements}
This work was supported in part by the National Natural Science Foundation of China under Grant 62125603, Grant 62321005, and Grant 62336004.

\appendix
\section*{Supplementary Material}
\noindent This supplementary material is organized as follows: 
\begin{itemize}
\itemsep0em 
    \item Section \ref{example} provides illustrated examples of our theoretical analysis.
    \item Section \ref{benchmark} details the ScanNet-md40 and TO-SCENE-down benchmarks. 
    \item Section \ref{category} details the experimental results with per-category APs. 
\end{itemize}
We also include a demo video and our source code in the supplementary material.

\section{Illustrated Examples}\label{example}
\begin{figure}[t]
  \centering
  \includegraphics[width=1.0\linewidth]{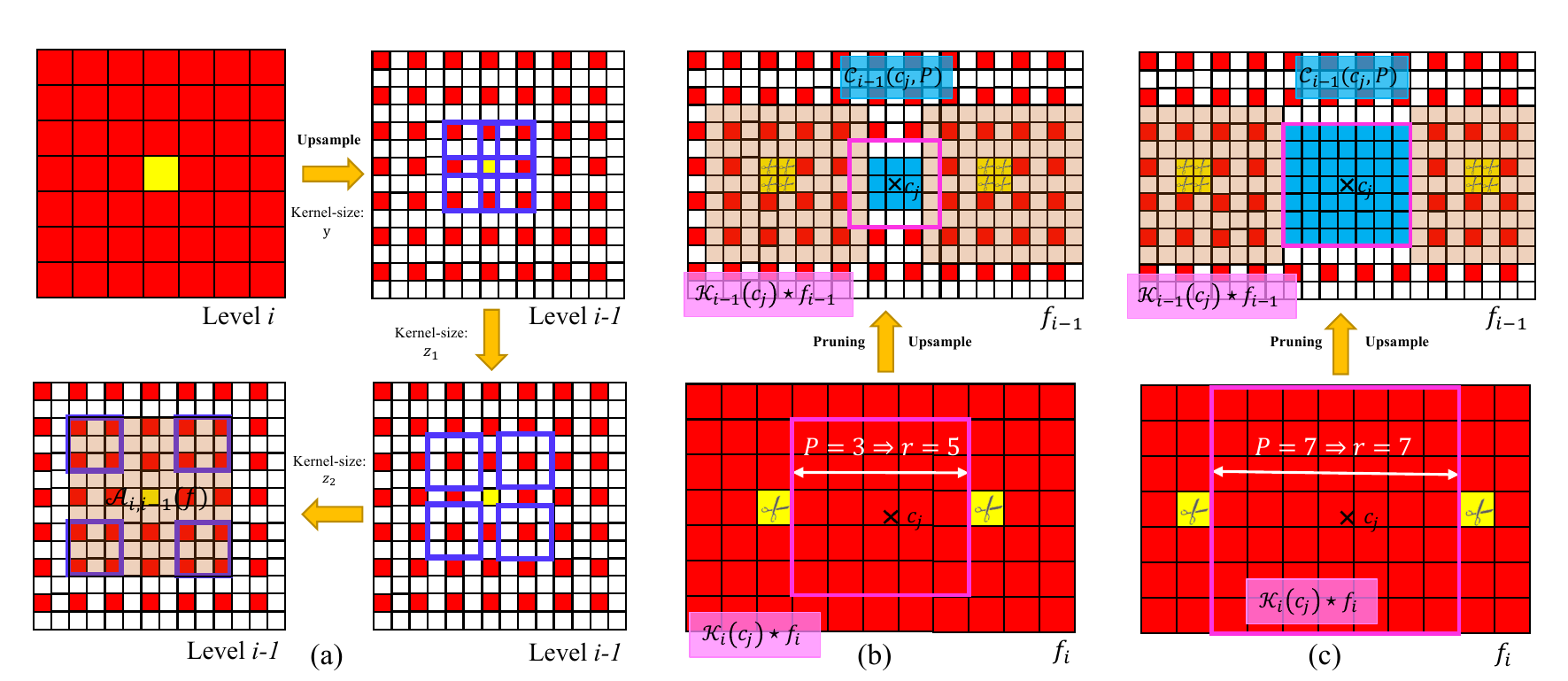}
  \caption{Visual demonstration of our pruning strategy. (a) shows how a voxel in level $i$ affects neighbor voxels in level $i-1$ after several convolutional layers. The affecting field $\mathcal{A}_{i,i-1}(f)$ is shown in translucent mask. (b) and (c) show how the choice of $P$ affects the pruning strategy, where the blue region is the unaffected $P\times P\times P$ voxels $\mathcal{C}_{i-1}(\boldsymbol{c_j}, P)$, the black cross is object center $c_j$ and the pink box is the pruning strategy $\mathcal{K}(c_j)$. All voxels outside the pink box will be pruned.}
  \label{diagram}
\end{figure}

We provide detailed explanation of Equation (1), (2), (3) and (4) in this section. Our theoretical derivation is shown in Figure \ref{diagram}.

\textbf{Eq1:} Assuming we are pruning $f_i$, without loss of generality, we only need to consider one object $c_j\ (j<i)$ to acquire $\mathcal{K}_i(c_j)$. Then we can summarize all $\mathcal{K}_i(c_j)$ to acquire the final pruning mask $M_i$.

\textbf{Eq2:} Figure \ref{diagram} (a) shows the expansion of affecting field between level $i$ and $i-1$. Eq (2) can be detailed as:
\begin{gather}
  aff(\{x_k\},y,\{z_k\})= \nonumber \\ 
  [1+\sum_{k=1}^m{(x_k-1)}]\cdot 2+1+y-1+\sum_{k=1}^n{(z_k-1)}
\end{gather}
Given $\{x_k\}=\varnothing$, $\{y\}=3$ and $\{z_k\}=\{3,3\}$, the affecting field $\mathcal{A}_{i,i-1}(v_i)$ is a cube of $9\times 9\times 9$ voxels in level $i-1$.

\textbf{Eq3:} According to Figure \ref{diagram} (b) and (c), to ensure $\mathcal{C}_j(\boldsymbol{c_j}, P)$ unaffected, we only need to ensure $\mathcal{C}_{i-1}(\boldsymbol{c_j}, P)$ unaffected, as we explained in Line 200-203, Page 7. With the help of (a), we can derive the range $r$ of pruning mask $\mathcal{K}_i(c_j)$, as shown in Eq (3). \textcolor{red}{In this way, we can prune all voxels outside the $r\times r\times r$ pink box in level $i$ without affecting $\mathcal{C}_{i-1}(\boldsymbol{c_j}, P)$.}

\textbf{Eq4:} However, we need to further ensure pruning in level $i$ has no cumulative affect on pruning in level $i-1$. When we prune $f_{i-1}$, if $j<i-1$, we still need to ensure $\mathcal{C}_{i-2}(\boldsymbol{c_j}, P)$ unaffected. Similarly, we can derive the pruning mask $\mathcal{K}_{i-1}(c_j)$, which is a cube with length $r\cdot S_{i-1}$. \textcolor{red}{Note that the unpruned voxels should be unaffected in order for the recursion to be correct. In another word, the pink box in level $i-1$ should be fully covered by the blue region.} This is the meaning of Eq (4). When $P=3$, then $r=5$, so that some unpruned voxels are still affected. When $P=7$, then $r=7$, which exactly ensures the unpruned voxels to be unaffected.

\begin{figure*}[t]
  \centering
  \includegraphics[width=1.0\linewidth]{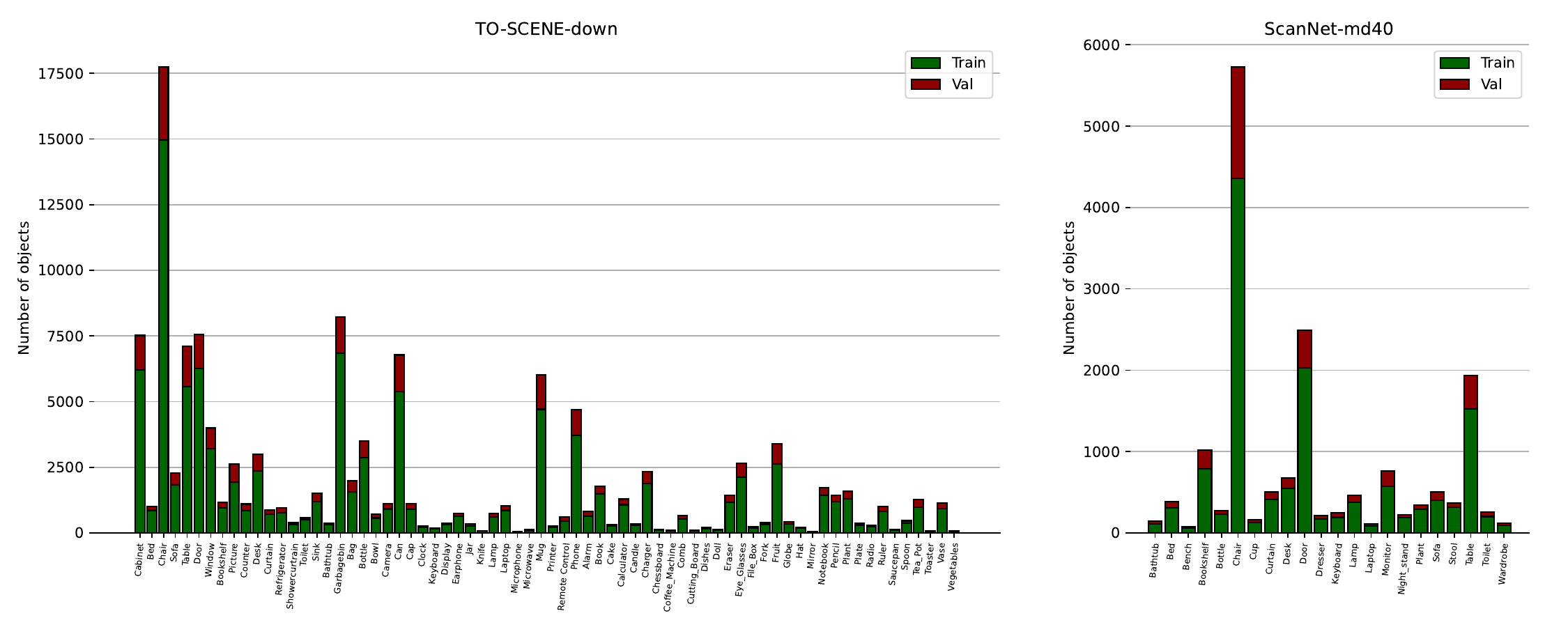}
  \caption{Number of objects in each category for ScanNet-md40 and TO-SCENE-down. For each category, we report the total number of objects on training and validation sets and use red and blue bars to distinguish the numbers on each set.}
  \label{num-two}
\end{figure*}

\section{Details on Benchmarks}\label{benchmark}
We demonstrate the detailed categories and the number of objects in each class for both benchmarks in Figure \ref{num-two}. We observe significant long tail effect in both benchmarks and find that the number of small objects in the real world dataset (i.e.\ ScanNet) is usually small, which poses great challenge to the 3D object detector.

\begin{figure}[t]
  \centering
  \includegraphics[width=0.7\linewidth]{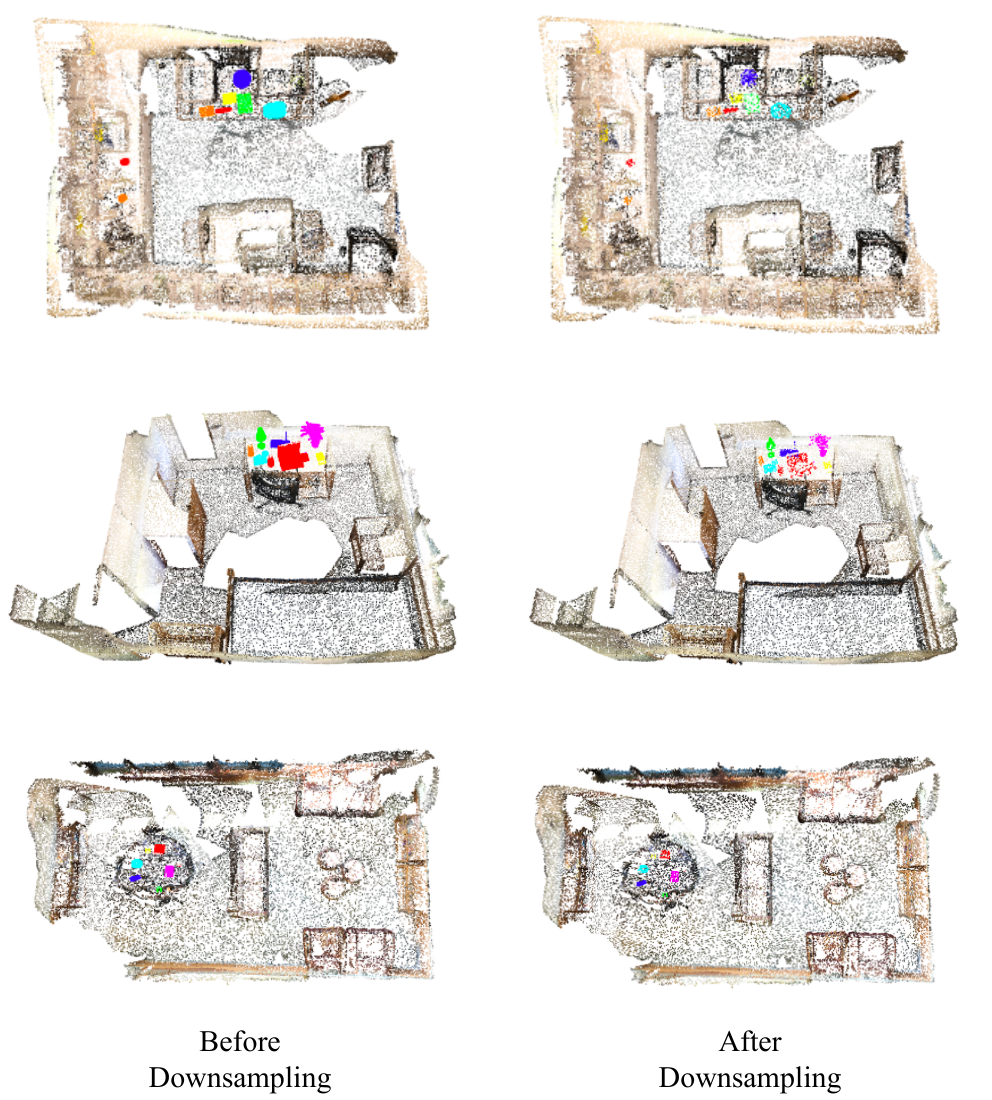}
  \caption{Visualization of scenes in TO-SCENE before (originally provided in \cite{xu2022scene}) and after (our TO-SCENE-down benchmark) downsampling. After downsampling, the density of points on small objects is closer to others, which is more realistic.}
  \label{compare}
\end{figure}

In order to prove the necessity of downsampling for small objects in TO-SCENE-down, we further visualize the scenes in TO-SCENE before and after downsampling in Figure \ref{compare}. It can be seen that the original dataset~\cite{xu2022scene} contains densely and non-uniformly sampled small objects, whose density is obviously much larger than other objects and backgrounds. While after our downsampling, the overall scenes are closer to naturally sampled real scenes.

\section{Class-specific Results}\label{category}
We provide more detailed experimental results on ScanNet-md40 and TO-SCENE-down with class-specific APs. Table \ref{tbl:scannet-percls-map25}, \ref{tbl:scannet-percls-map50}, \ref{tbl:toscene-percls-map25} and \ref{tbl:toscene-percls-map50} refer to AP@0.25 on ScanNet-md40, AP@0.5 on ScanNet-md40, AP@0.25 on TO-SCENE-down and AP@0.5 on TO-SCENE-down respectively.
We highlight the categories of small objects in blue. It can be seen that DSPDet3D achieves much better performance on small objects compared with the state-of-the-arts.

\begin{table*}[t]
  \centering
  \footnotesize
  \caption{The class-specific detection results (AP@0.25) of different methods on ScanNet-md40 benchmark. We highlight the categories of small objects in blue.}
  \vspace{2mm}
    \begin{tabular}{c|ccccccccccc|cc}  
      \toprule
       & \rotatebox{75}{VoteNet} & \rotatebox{75}{VoteNet$_S$} & \rotatebox{75}{H3DNet} & \rotatebox{75}{GroupFree3D} & \rotatebox{75}{GroupFree$_S$} & \rotatebox{75}{RBGNet} & \rotatebox{75}{CAGroup3D} & \rotatebox{75}{FCAF3D} & \rotatebox{75}{TR3D} & \rotatebox{75}{FCAF3D\_higher} & \rotatebox{75}{TR3D\_higher} & \rotatebox{75}{Ours($\tau=0$)} & \rotatebox{75}{\textbf{Ours($\boldsymbol{\tau=0.3}$)}} \\
      \midrule
    Bathtub & 90.39 & 84.26 & 91.46 & 92.70 & 71.36 & 44.22 & 44.69 & 83.65 & 93.93 & 88.16 &84.89 & 86.77 & 86.77 \\
    Bed   & 87.30 & 89.03 & 88.46 & 89.11 & 71.58 & 90.74 & 87.74 & 86.91 & 87.64 &88.28 &87.01 & 89.31 & 89.38 \\
    Bench & 48.34 & 40.56 & 46.89 & 38.20 & 10.01 & 47.79 & 42.09 & 39.20 & 53.05 &48.73 &53.58 & 52.93 & 53.15 \\
    Bookshelf & 55.47 & 54.43 & 59.88 & 58.90 & 20.97 & 61.86 & 69.09 & 65.88 & 66.17 &66.01 &62.27 & 67.06 & 67.43 \\
    \rowcolor{myhighlight}Bottle & 0.00  & 0.06  & 1.22  & 4.16  & 0.00  & 0.20  & 5.56  & 2.98  & 3.23 &12.19 &26.58  & 26.91 & 17.36 \\
    Chair & 89.03 & 89.15 & 91.70 & 93.20 & 56.76 & 94.16 & 95.04 & 94.96 & 95.58 &94.90 &95.52 & 95.68 & 95.65 \\
    \rowcolor{myhighlight}Cup   & 0.00  & 0.00  & 0.00  & 4.37  & 0.01  & 0.00  & 0.00  & 5.85  & 12.95 &6.60 &14.60 & 20.86 & 24.04 \\
    Curtain & 56.69 & 59.17 & 67.00 & 70.31 & 27.31 & 62.81 & 72.45 & 68.00 & 62.54 &67.06 &69.52 & 70.73 & 70.72 \\
    Desk  & 68.81 & 60.05 & 76.48 & 75.13 & 40.43 & 76.83 & 79.21 & 74.94 & 74.27 &70.04 &74.33 & 72.83 & 72.87 \\
    Door  & 53.17 & 48.57 & 59.45 & 61.62 & 16.70 & 59.55 & 61.78 & 60.26 & 60.91 &63.61 &65.06 & 67.68 & 67.82 \\
    Dresser & 32.63 & 38.88 & 34.70 & 48.60 & 28.88 & 40.23 & 60.42 & 31.62 & 40.82 &37.22 &36.81 & 28.67 & 26.37 \\
    \rowcolor{myhighlight}Keyboard & 0.03  & 0.19  & 0.09  & 3.19  & 0.00  & 0.00  & 22.37 & 27.96 & 46.58 &44.27 &58.24 & 58.55 & 60.91 \\
    Lamp  & 49.26 & 43.29 & 49.33 & 59.05 & 8.76  & 56.90 & 61.19 & 61.00 & 61.32 &65.55 &62.65 & 64.34 & 64.60 \\
    \rowcolor{myhighlight}Laptop & 1.17  & 3.91  & 11.09 & 19.10 & 0.81  & 23.11 & 38.45 & 36.95 & 47.31 &47.64 &67.38 & 72.85 & 72.77 \\
    Monitor & 67.18 & 68.76 & 74.88 & 83.35 & 30.12 & 83.63 & 86.93 & 88.71 & 89.20 &91.73 &88.96 & 92.87 & 91.49 \\
    Night\_Stand & 81.35 & 78.26 & 86.12 & 83.77 & 56.51 & 80.86 & 91.62 & 90.39 & 91.44 &80.64 &92.10 & 81.95 & 81.41 \\
    Plant & 29.36 & 18.08 & 29.45 & 21.71 & 4.24  & 50.47 & 59.87 & 43.29 & 35.78 &61.35 &58.18 & 65.92 & 63.95 \\
    Sofa  & 88.89 & 87.33 & 89.09 & 84.79 & 53.99 & 91.32 & 91.41 & 88.38 & 91.32 &85.49 &89.90 & 90.62 & 90.61 \\
    Stool & 44.34 & 37.72 & 37.63 & 44.54 & 17.66 & 51.41 & 55.25 & 52.51 & 42.65 &53.21 &46.76 & 43.95 & 45.49 \\
    Table & 64.35 & 60.42 & 65.69 & 74.31 & 32.61 & 74.41 & 70.37 & 71.47 & 69.99 &69.52 &70.46 & 70.66 & 70.63 \\
    Toilet & 94.01 & 95.48 & 97.89 & 95.82 & 82.95 & 97.76 & 99.73 & 99.39 & 98.81 &99.83 &99.60 & 99.85 & 99.85 \\
    Wardrobe & 20.67 & 12.04 & 18.72 & 43.01 & 16.02 & 26.80 & 31.12 & 34.48 & 29.49 &36.17 &29.51 & 17.61 & 17.66 \\
      \bottomrule
    \end{tabular}
  \label{tbl:scannet-percls-map25}
\end{table*}

\begin{table*}[t]
  \centering
  \footnotesize
  \caption{The class-specific detection results (AP@0.5) of different methods on ScanNet-md40 benchmark. We highlight the categories of small objects in blue.}
  \vspace{2mm}
    \begin{tabular}{c|ccccccccccc|cc}  
      \toprule
       & \rotatebox{75}{VoteNet} & \rotatebox{75}{VoteNet$_S$} & \rotatebox{75}{H3DNet} & \rotatebox{75}{GroupFree3D} & \rotatebox{75}{GroupFree$_S$} & \rotatebox{75}{RBGNet} & \rotatebox{75}{CAGroup3D} & \rotatebox{75}{FCAF3D} & \rotatebox{75}{TR3D} & \rotatebox{75}{FCAF3D\_higher} & \rotatebox{75}{TR3D\_higher} & \rotatebox{75}{Ours($\tau=0$)} & \rotatebox{75}{\textbf{Ours($\boldsymbol{\tau=0.3}$)}} \\
      \midrule
      Bathtub & 79.13 & 84.26 & 86.16 & 85.47 & 44.68 & 39.46 & 42.97 & 83.65 & 84.21 &77.90 &78.36 & 81.07 & 81.07 \\
    Bed   & 82.43 & 80.35 & 82.69 & 81.85 & 42.71 & 78.17 & 81.93 & 80.91 & 82.32 &82.31 &80.86 & 83.47 & 83.50 \\
    Bench & 1.63  & 4.57  & 22.45 & 2.36  & 0.51  & 1.53  & 14.50 & 27.95 & 34.15 &27.19 &43.74 & 41.68 & 41.79 \\
    Bookshelf & 34.16 & 29.23 & 43.99 & 44.51 & 4.07  & 44.06 & 55.92 & 56.06 & 54.22 &56.38 &51.46 & 56.55 & 56.84 \\
    \rowcolor{myhighlight}Bottle & 0.00  & 0.00  & 0.00  & 0.08  & 0.00  & 0.00  & 1.44  & 0.48  & 1.50  &8.47 &19.09 & 20.91 & 13.40 \\
    Chair & 74.22 & 69.87 & 80.52 & 83.41 & 14.31 & 82.33 & 90.19 & 89.98 & 91.13 &90.01 &91.53 & 91.36 & 91.28 \\
    \rowcolor{myhighlight}Cup   & 0.00  & 0.00  & 0.00  & 0.00  & 0.00  & 0.00  & 0.00  & 2.69  & 5.98  &5.74 &13.94 & 18.81 & 21.51 \\
    Curtain & 19.27 & 17.46 & 30.86 & 44.91 & 1.16  & 15.56 & 44.52 & 43.62 & 34.97 &40.01 &42.53 & 44.51 & 44.51 \\
    Desk  & 36.27 & 32.54 & 49.13 & 47.80 & 10.95 & 45.39 & 60.25 & 54.36 & 58.66 &50.36 &59.03 & 56.36 & 56.37 \\
    Door  & 22.37 & 18.54 & 32.57 & 38.00 & 2.05  & 34.14 & 46.36 & 42.34 & 43.72 &42.28 &48.50 & 49.40 & 49.54 \\
    Dresser & 22.24 & 23.72 & 26.66 & 35.84 & 4.80  & 24.53 & 47.19 & 25.14 & 35.14 &29.50 &30.08 & 23.04 & 20.89 \\
    \rowcolor{myhighlight}Keyboard & 0.00  & 0.00  & 0.00  & 0.84  & 0.00  & 0.00  & 5.40  & 2.89  & 13.36 &25.61 &38.15& 37.87 & 36.47 \\
    Lamp  & 21.95 & 11.31 & 28.96 & 35.29 & 0.08  & 26.10 & 53.80 & 47.10 & 42.60 &51.48 &44.77 & 49.65 & 50.62 \\
    \rowcolor{myhighlight}Laptop & 0.00  & 0.04  & 3.60  & 2.30  & 0.00  & 0.00  & 27.37 & 26.77 & 29.47 &25.10 &47.07 & 48.61 & 50.12 \\
    Monitor & 28.18 & 24.70 & 35.05 & 42.82 & 2.01  & 30.24 & 68.77 & 66.31 & 74.31 &73.87 &74.63 & 75.01 & 76.03 \\
    Night\_Stand & 70.94 & 59.96 & 78.42 & 71.77 & 31.82 & 71.30 & 91.62 & 83.49 & 89.86 &80.15 &86.26 & 81.46 & 80.15 \\
    Plant & 11.59 & 9.82  & 20.37 & 13.21 & 0.44  & 26.55 & 52.90 & 35.56 & 30.17 &52.48 &52.00 & 54.07 & 53.41 \\
    Sofa  & 72.68 & 76.60 & 74.89 & 70.23 & 24.38 & 67.25 & 83.29 & 78.04 & 78.14 &77.04 &83.63 & 83.00 & 82.99 \\
    Stool & 28.24 & 21.08 & 25.53 & 40.49 & 7.10  & 15.68 & 51.59 & 47.54 & 40.50 &46.59 &43.85 & 40.99 & 42.27 \\
    Table & 48.72 & 39.26 & 47.40 & 59.83 & 9.22  & 38.50 & 61.56 & 59.82 & 60.41 &59.87 &58.51& 59.28 & 59.20 \\
    Toilet & 81.74 & 86.43 & 85.06 & 89.99 & 61.14 & 75.64 & 96.08 & 95.72 & 87.51 &97.65 &90.95& 95.98 & 96.07 \\
    Wardrobe & 5.42  & 4.36  & 8.75  & 19.58 & 1.25  & 1.65  & 20.15 & 22.08 & 27.23 &22.31 &9.63& 8.03  & 7.58 \\
      \bottomrule
    \end{tabular}
  \label{tbl:scannet-percls-map50}
\end{table*}

\begin{table*}[t]
  \centering
  \tiny
  \caption{The class-specific detection results (AP@0.25) of different methods on To-Scene benchmark. We highlight the categories of small objects in blue.}
  \vspace{2mm}
    \begin{tabular}{c|ccccccccccc|cc}  
      \toprule
       & \rotatebox{75}{VoteNet} & \rotatebox{75}{VoteNet$_S$} & \rotatebox{75}{H3DNet} & \rotatebox{75}{GroupFree3D} & \rotatebox{75}{GroupFree$_S$} & \rotatebox{75}{RBGNet} & \rotatebox{75}{CAGroup3D} & \rotatebox{75}{FCAF3D} & \rotatebox{75}{TR3D} & \rotatebox{75}{FCAF3D\_higher} & \rotatebox{75}{TR3D\_higher} & \rotatebox{75}{Ours($\tau=0$)} & \rotatebox{75}{\textbf{Ours($\boldsymbol{\tau=0.5}$)}} \\
      \midrule
    Cabinet & 53.34 & 49.72 & 58.94 & 61.90 & 57.12 & 65.36 & 66.65 & 63.93 & 61.56 &67.44 &67.26  &64.05 & 64.36 \\
    Bed   & 84.96 & 82.54 & 80.13 & 82.13 & 80.25 & 92.11 & 84.66 & 80.93 & 78.29 &75.81 &71.90 & 76.28 & 76.28 \\
    Chair & 87.79 & 85.54 & 89.54 & 92.55 & 88.29 & 92.57 & 94.20 & 92.70 & 94.14 &93.65 &93.97 & 93.48 & 93.61 \\
    Sofa  & 92.00 & 89.87 & 87.43 & 91.32 & 86.33 & 88.96 & 86.99 & 89.75 & 90.65 
    &87.81 &90.11 & 91.62 & 91.65 \\
    Table & 67.19 & 64.44 & 68.24 & 74.24 & 69.97 & 86.21 & 73.25 & 75.72 & 79.42 &78.25 &77.30 & 78.99 & 78.99 \\
    Door  & 51.39 & 52.88 & 53.85 & 60.60 & 53.62 & 59.99 & 59.99 & 50.37 & 55.88 &58.44 &58.68 & 57.40 & 57.05 \\
    Window & 41.05 & 43.40 & 44.87 & 48.21 & 41.62 & 58.83 & 53.19 & 46.33 & 43.79 &47.99 &45.21 & 43.10 & 41.56 \\
    Bookshelf  & 29.69 & 24.08 & 30.24 & 28.43 & 27.74 & 27.52 & 26.29 & 21.24 & 29.02 &32.48 &24.57 & 34.26 & 34.66 \\
    Picture & 6.60  & 6.78  & 8.26  & 11.73 & 8.28  & 28.63 & 26.59 & 14.04 & 11.80 &18.73 &21.17 & 20.34 & 17.01 \\
    Counter & 56.73 & 51.35 & 63.49 & 62.73 & 63.39 & 71.11 & 71.83 & 67.90 & 63.98 &62.67 &65.65 & 58.33 & 59.14 \\
    Desk  & 59.65 & 57.20 & 56.39 & 59.23 & 49.80 & 64.55 & 64.29 & 63.38 & 59.20 &61.12 &62.24 & 59.03 & 59.30 \\
    Curtain & 48.39 & 46.38 & 54.85 & 54.23 & 56.41 & 37.68 & 65.68 & 54.77 & 43.18 &51.79 &53.61 & 54.46 & 54.64 \\
    Refrigerator & 49.68 & 54.76 & 69.77 & 65.12 & 41.31 & 67.51 & 74.50 & 75.35 & 65.85 &68.11 &64.06 & 71.38 & 71.80 \\
    Showercurtrain & 75.65 & 78.95 & 75.76 & 78.47 & 81.55 & 83.46 & 79.70 & 78.48 & 39.93 &82.46 &56.37 & 57.29 & 51.09 \\
    Toilet & 99.73 & 99.98 & 100.00 & 100.00 & 100.00 & 100.00 & 99.58 & 100.00 & 100.00 &100.00 &100.00 & 100.00 & 100.00 \\
    Sink  & 66.61 & 73.02 & 85.24 & 85.41 & 84.65 & 84.52 & 85.49 & 90.04 & 92.18 &92.30 &89.94 & 88.11 & 87.58 \\
    Bathtub & 95.80 & 95.42 & 95.74 & 95.91 & 96.26 & 94.54 & 96.02 & 93.14 & 94.80 &96.30 &95.89 & 96.30 & 96.30 \\
    Garbagebin & 51.42 & 49.21 & 56.03 & 68.63 & 58.57 & 68.96 & 69.86 & 68.51 & 66.30 &71.44 &71.36 & 72.83 & 70.77 \\
    \rowcolor{myhighlight}Bag   & 39.22 & 49.08 & 44.83 & 60.35 & 69.97 & 73.75 & 89.27 & 83.63 & 89.28 &89.33 &91.71 & 91.90 & 91.19 \\
    \rowcolor{myhighlight}Bottle & 15.00 & 30.67 & 20.20 & 33.24 & 53.09 & 58.26 & 69.44 & 35.88 & 70.83 &68.38 &83.14 & 83.06 & 83.31 \\
    \rowcolor{myhighlight}Bowl  & 11.79 & 23.27 & 8.72  & 12.58 & 44.17 & 39.89 & 85.21 & 55.08 & 73.95 &85.52 &93.08 & 92.29 & 92.71 \\
    \rowcolor{myhighlight}Camera & 7.88  & 15.57 & 9.40  & 6.91  & 24.00 & 13.06 & 53.30 & 44.11 & 62.24 &65.23 &70.69 & 79.41 & 78.45 \\
    \rowcolor{myhighlight}Can   & 12.40 & 22.05 & 14.65 & 26.08 & 41.43 & 45.35 & 76.29 & 42.40 & 76.81 &70.54 &89.41 & 89.39 & 89.63 \\
    \rowcolor{myhighlight}Cap   & 19.46 & 46.49 & 15.53 & 34.34 & 60.86 & 45.89 & 84.58 & 67.04 & 82.38 &78.98 &86.72 & 92.02 & 91.90 \\
    \rowcolor{myhighlight}Clock & 1.29  & 1.87  & 0.94  & 3.56  & 5.42  & 14.13 & 11.30 & 2.65  & 26.69 &23.17 &30.45 & 36.12 & 36.83 \\
    \rowcolor{myhighlight}Keyboard & 0.14  & 14.35 & 3.05  & 5.20  & 26.71 & 0.86  & 56.43 & 38.04 & 76.35 &54.45 &91.91 & 93.94 & 90.51 \\
    \rowcolor{myhighlight}Display & 53.27 & 47.56 & 54.15 & 46.78 & 66.93 & 81.61 & 88.27 & 86.55 & 87.09 &81.23 &92.96 & 93.94 & 93.96 \\
    \rowcolor{myhighlight}Earphone & 8.94  & 28.85 & 6.06  & 17.31 & 46.37 & 27.75 & 73.42 & 56.00 & 75.10 &64.67 &80.53 & 81.71 & 81.40 \\
    \rowcolor{myhighlight}Jar  & 5.31  & 29.54 & 6.75  & 7.75  & 28.42 & 25.47 & 34.58 & 25.63 & 23.92 &38.75 &39.66 & 33.20 & 33.18 \\
    \rowcolor{myhighlight}Knife & 0.63  & 0.00  & 0.18  & 1.72  & 1.26  & 1.26  & 0.00  & 0.24  & 16.52 &18.48 &32.73 & 31.68 & 31.70 \\
    \rowcolor{myhighlight}Lamp  & 34.45 & 53.44 & 41.38 & 58.99 & 67.95 & 72.67 & 89.10 & 78.45 & 83.13 &90.53 &89.47 & 92.34 & 92.16 \\
    \rowcolor{myhighlight}Laptop & 65.25 & 66.98 & 67.75 & 82.17 & 94.18 & 88.77 & 96.86 & 95.94 & 96.83 &97.93 &98.15 & 98.00 & 97.12 \\
    \rowcolor{myhighlight}Microphone & 0.04  & 0.00  & 0.01  & 0.01  & 0.04  & 0.16  & 0.00  & 0.16  & 0.96  &7.87 &1.38 & 2.26  & 2.45 \\
    \rowcolor{myhighlight}Microwave & 50.07 & 52.79 & 50.18 & 54.93 & 52.49 & 63.78 & 82.98 & 78.13 & 69.59 &89.10 &84.96 & 90.68 & 90.01 \\
    \rowcolor{myhighlight}Mug   & 13.90 & 29.48 & 15.80 & 24.09 & 48.59 & 39.16 & 85.77 & 53.40 & 78.14 &76.78 &93.69 & 94.18 & 93.86 \\
    \rowcolor{myhighlight}Printer  & 27.09 & 22.25 & 42.07 & 43.61 & 35.57 & 54.19 & 65.72 & 68.49 & 66.63 &70.62 &73.67 & 65.19 & 64.06 \\
    \rowcolor{myhighlight}Remote Control & 0.36  & 2.33  & 0.22  & 0.32  & 2.00  & 2.68  & 12.38 & 1.68  & 34.58 &17.03 &50.49 & 60.44 & 60.42 \\
    \rowcolor{myhighlight}Phone & 1.52  & 6.22  & 1.61  & 2.89  & 15.96 & 14.54 & 29.13 & 9.97  & 66.23 &29.49 &79.46 & 84.59 & 84.60 \\
    \rowcolor{myhighlight}Alarm & 3.07  & 12.11 & 3.56  & 9.51  & 19.04 & 13.44 & 39.08 & 20.60 & 30.59 &43.89 &48.12 & 56.57 & 58.09 \\
    \rowcolor{myhighlight}Book  & 20.37 & 31.58 & 27.02 & 31.09 & 38.88 & 33.74 & 57.75 & 34.95 & 63.88 &59.24 &72.07 & 76.00 & 76.61 \\
    \rowcolor{myhighlight}Cake  & 20.69 & 27.18 & 22.09 & 27.41 & 38.48 & 31.28 & 64.14 & 56.94 & 54.51 &65.54 &56.40 & 65.61 & 64.76 \\
    \rowcolor{myhighlight}Calculator & 1.51  & 6.34  & 1.99  & 2.88  & 13.71 & 11.74 & 21.56 & 16.73 & 34.34 &31.16 &47.77 & 53.43 & 52.72 \\
    \rowcolor{myhighlight}Candle & 28.00 & 29.63 & 21.62 & 42.58 & 49.31 & 53.19 & 56.87 & 41.87 & 65.45 &71.23 &70.38 & 69.32 & 69.76 \\
    \rowcolor{myhighlight}Charger & 0.03  & 2.12  & 0.33  & 0.53  & 1.78  & 8.09  & 22.22 & 6.33  & 37.47 &34.47 &58.48 & 65.44 & 62.81 \\
    \rowcolor{myhighlight}Chessboard & 6.80  & 45.14 & 19.33 & 27.38 & 74.76 & 71.96 & 87.45 & 78.31 & 77.94 &73.06 &82.03 & 89.12 & 89.72 \\
    \rowcolor{myhighlight}Coffee\_Machine & 41.21 & 27.28 & 32.04 & 34.88 & 47.09 & 53.52 & 77.94 & 62.97 & 38.69 &56.74 &49.34 & 60.68 & 60.85 \\
    \rowcolor{myhighlight}Comb  & 0.30  & 1.21  & 0.12  & 1.67  & 6.28  & 4.82  & 11.05 & 2.44  & 23.96 &34.23 &48.43 & 53.38 & 45.74 \\
    \rowcolor{myhighlight}Cutting\_Board & 10.57 & 8.29  & 14.22 & 17.90 & 38.76 & 30.21 & 0.00  & 32.18 & 65.59 &77.53 &72.34 & 69.72 & 62.20 \\
    \rowcolor{myhighlight}Dishes & 11.25 & 26.30 & 9.16  & 21.03 & 40.12 & 26.11 & 70.50 & 42.89 & 64.15 &77.43 &72.34 & 71.50 & 68.77 \\
    \rowcolor{myhighlight}Doll  & 1.14  & 1.89  & 0.70  & 7.24  & 17.74 & 2.12  & 9.55  & 1.68  & 14.54 &15.49 &10.21 & 28.84 & 25.06 \\
    \rowcolor{myhighlight}Eraser & 0.00  & 0.04  & 0.00  & 0.00  & 0.00  & 0.55  & 0.00  & 0.29  & 34.87 &13.71 &61.28 & 66.13 & 60.98 \\
    \rowcolor{myhighlight}Eye\_Glasses & 5.67  & 23.20 & 7.84  & 12.57 & 40.69 & 31.52 & 81.29 & 58.37 & 91.95 &76.77 &97.20 & 98.20 & 98.41 \\
    \rowcolor{myhighlight}File\_Box & 56.97 & 49.39 & 33.74 & 40.07 & 45.78 & 55.42 & 63.47 & 60.84 & 67.27 &70.03 &73.19 & 71.86 & 72.55 \\
    \rowcolor{myhighlight}Fork  & 0.84  & 0.61  & 1.76  & 1.31  & 1.86  & 1.49  & 6.52  & 6.84  & 17.14 &13.33 &29.90 & 30.95 & 32.48 \\
    \rowcolor{myhighlight}Fruit & 2.56  & 9.54  & 2.15  & 7.53  & 29.77 & 20.58 & 62.04 & 32.09 & 52.27 &62.27 &80.03 & 81.95 & 82.03 \\
    \rowcolor{myhighlight}Globe & 30.87 & 29.41 & 19.91 & 39.84 & 52.62 & 35.67 & 75.65 & 64.64 & 65.02 &74.12 &80.54 & 79.33 & 78.30 \\
    \rowcolor{myhighlight}Hat   & 1.87  & 22.49 & 2.95  & 4.18  & 13.18 & 6.83  & 53.65 & 23.64 & 32.08 &40.68 &50.14 & 52.34 & 49.58 \\
    \rowcolor{myhighlight}Mirror & 0.70  & 17.36 & 0.76  & 0.20  & 0.54  & 5.63  & 0.58  & 28.93 & 1.30 &23.89 &1.55  & 3.25  & 2.87 \\
    \rowcolor{myhighlight}Notebook & 3.90  & 8.12  & 8.52  & 8.33  & 23.19 & 26.65 & 37.44 & 17.61 & 59.38 &30.12 &63.50 & 69.06 & 68.64 \\
    \rowcolor{myhighlight}Pencil & 0.01  & 1.14  & 0.08  & 0.06  & 1.00  & 1.51  & 0.00  & 3.47  & 32.09 &28.53 &55.31 & 65.04 & 65.32 \\
    \rowcolor{myhighlight}Plant & 31.59 & 49.74 & 35.44 & 53.65 & 65.60 & 60.77 & 87.03 & 73.41 & 78.99 &86.43 &88.62 & 87.34 & 87.54 \\
    \rowcolor{myhighlight}Plate & 5.31  & 20.83 & 7.71  & 10.31 & 27.52 & 38.64 & 60.91 & 19.40 & 80.18 &63.41 &96.94 & 98.03 & 98.03 \\
    \rowcolor{myhighlight}Radio & 1.95  & 5.50  & 5.59  & 3.13  & 6.80  & 12.77 & 15.56 & 10.42 & 18.59 &25.11 &25.87 & 32.59 & 32.35 \\
    \rowcolor{myhighlight}Ruler & 0.02  & 1.25  & 0.42  & 0.13  & 1.68  & 0.96  & 3.99  & 1.09  & 35.56 &14.91 &58.44 & 61.15 & 61.78 \\
    \rowcolor{myhighlight}Saucepan & 31.34 & 32.74 & 24.73 & 39.52 & 47.05 & 37.65 & 73.89 & 37.29 & 30.28 &48.26 &50.70 & 63.28 & 63.36 \\
    \rowcolor{myhighlight}Spoon & 0.05  & 1.76  & 1.19  & 0.65  & 4.04  & 5.97  & 9.89  & 2.90  & 27.35 &29.13 &33.52 & 38.93 & 38.14 \\
    \rowcolor{myhighlight}Tea\_Pot & 21.10 & 28.32 & 17.05 & 33.06 & 48.87 & 42.51 & 89.43 & 76.64 & 77.67 &89.80 &88.35 & 91.76 & 91.92 \\
    \rowcolor{myhighlight}Toaster & 13.88 & 21.95 & 19.51 & 11.42 & 16.27 & 19.90 & 34.61 & 31.32 & 29.11 &37.50 &24.48 & 34.88 & 35.01 \\
    \rowcolor{myhighlight}Vase & 25.95 & 37.29 & 17.00 & 31.59 & 46.84 & 37.72 & 65.79 & 52.50 & 54.39 &64.01 &67.67 & 73.63 & 73.27 \\
    \rowcolor{myhighlight}Vegetables & 0.43  & 7.12  & 0.40  & 0.66  & 18.35 & 10.27 & 0.01  & 4.73  & 9.62  &9.79 &7.06 & 14.00  & 13.52 \\
      \bottomrule
    \end{tabular}
  \label{tbl:toscene-percls-map25}
\end{table*}

\begin{table*}[t]
  \centering
  \tiny
  \caption{The class-specific detection results (AP@0.5) of different methods on To-Scene benchmark. We highlight the categories of small objects in blue.}
  \vspace{2mm}
    \begin{tabular}{c|ccccccccccc|cc}  
      \toprule
       & \rotatebox{75}{VoteNet} & \rotatebox{75}{VoteNet$_S$} & \rotatebox{75}{H3DNet} & \rotatebox{75}{GroupFree3D} & \rotatebox{75}{GroupFree$_S$} & \rotatebox{75}{RBGNet} & \rotatebox{75}{CAGroup3D} & \rotatebox{75}{FCAF3D} & \rotatebox{75}{TR3D} & \rotatebox{75}{FCAF3D\_higher} & \rotatebox{75}{TR3D\_higher}& \rotatebox{75}{Ours($\tau=0$)} & \rotatebox{75}{\textbf{Ours($\boldsymbol{\tau=0.5}$)}} \\
      \midrule
      Cabinet & 19.75 & 18.39 & 25.36 & 34.62 & 25.64 & 40.75 & 48.45 & 36.36 & 37.93 &46.04 &45.36 & 42.75 & 42.93 \\
    Bed   & 73.12 & 76.46 & 78.68 & 74.08 & 71.35 & 90.37 & 74.66 & 75.89 & 67.55 &69.50 &64.60 & 67.20 & 67.21 \\
    Chair & 71.68 & 66.72 & 79.05 & 84.35 & 77.56 & 85.99 & 90.35 & 87.31 & 89.64 &88.29 &90.28 & 88.48 & 88.34 \\
    Sofa  & 86.54 & 82.29 & 84.88 & 86.89 & 79.51 & 83.64 & 81.79 & 88.81 & 90.65 &78.45 &88.87 & 88.61 & 88.34 \\
    Table & 51.04 & 47.21 & 58.03 & 66.25 & 59.27 & 76.61 & 69.99 & 71.10 & 72.52 &72.11 &70.76 & 72.76 & 72.78 \\
    Door  & 25.72 & 24.80 & 33.47 & 43.78 & 33.32 & 41.23 & 49.61 & 39.67 & 42.73 &45.75 &46.82 & 44.34 & 43.53 \\
    Window & 18.49 & 188.00 & 20.23 & 22.56 & 18.05 & 34.76 & 31.92 & 24.36 & 24.19 &23.29 &21.30 & 15.27 & 15.48 \\
    Bookshelf  & 22.22 & 14.42 & 15.49 & 12.83 & 16.29 & 20.03 & 22.90 & 10.33 & 22.22 &23.78 &20.23 & 30.47 & 30.81 \\
    Picture & 0.69  & 0.35  & 1.75  & 6.88  & 3.42  & 13.88 & 13.48 & 5.12  & 3.50  &13.51 &19.41 & 15.87 & 8.27 \\
    Counter & 18.08 & 9.07  & 17.45 & 37.21 & 31.22 & 40.43 & 49.28 & 43.81 & 40.71 &45.46 &31.81 & 29.12 & 29.43 \\
    Desk  & 35.78 & 26.70 & 35.47 & 45.83 & 37.67 & 45.68 & 37.99 & 42.25 & 42.13 &33.71 &40.19 & 39.06 & 39.18 \\
    Curtain & 11.85 & 13.01 & 6.57  & 15.00 & 19.05 & 12.10 & 19.58 & 10.60 & 15.42 &9.24 &14.46 & 21.27 & 21.37 \\
    Refrigerator & 38.74 & 33.15 & 47.15 & 27.88 & 21.22 & 53.58 & 64.15 & 54.13 & 50.53 &59.20 &62.29 & 57.12 & 57.40 \\
    Showercurtrain & 41.76 & 41.14 & 70.98 & 71.58 & 40.11 & 78.05 & 68.52 & 72.77 & 30.12 &73.76 &44.63 & 52.55 & 46.85 \\
    Toilet & 92.55 & 92.28 & 100.00 & 100.00 & 100.00 & 100.00 & 99.58 & 100.00 & 100.00 & 100.00 & 100.00 & 100.00 & 100.00\\
    Sink  & 19.97 & 22.51 & 38.04 & 43.85 & 42.29 & 40.28 & 51.54 & 46.37 & 55.16 &53.83 &44.04 & 52.09 & 50.92 \\
    Bathtub & 72.46 & 81.61 & 79.25 & 80.75 & 80.94 & 81.01 & 81.27 & 88.77 & 85.38 &90.30 &83.50 & 76.03 & 76.03 \\
    Garbagebin & 27.98 & 23.46 & 40.25 & 50.67 & 41.58 & 56.65 & 64.57 & 60.39 & 57.68 &65.61 &64.48 & 63.97 & 61.21 \\
    \rowcolor{myhighlight}Bag   & 12.48 & 18.08 & 20.08 & 26.86 & 41.31 & 57.46 & 82.31 & 75.13 & 81.16 &87.70 &87.94 & 89.19 & 88.55 \\
    \rowcolor{myhighlight}Bottle & 3.35  & 7.77  & 6.43  & 8.22  & 13.78 & 43.00 & 66.27 & 27.81 & 65.95 &66.51 &64.48 & 82.36 & 82.31 \\
    \rowcolor{myhighlight}Bowl  & 0.81  & 5.34  & 0.50  & 0.49  & 12.83 & 19.49 & 82.56 & 52.06 & 68.56 &82.05 &92.50 & 92.29 & 92.71 \\
    \rowcolor{myhighlight}Camera & 2.19  & 2.05  & 3.38  & 1.52  & 10.81 & 8.98  & 49.93 & 35.38 & 57.19 &64.43 &68.06 & 75.50 & 74.32 \\
    \rowcolor{myhighlight}Can   & 2.06  & 3.50  & 3.60  & 5.47  & 10.44 & 36.19 & 73.51 & 31.74 & 68.29 &66.53 &86.73 & 87.88 & 88.00 \\
    \rowcolor{myhighlight}Cap   & 1.69  & 16.38 & 6.49  & 10.87 & 33.43 & 37.72 & 80.34 & 64.26 & 82.36 &78.63 &86.67 & 91.81 & 91.36 \\
    \rowcolor{myhighlight}Clock & 0.00  & 0.19  & 0.00  & 2.73  & 0.40  & 8.94  & 8.68  & 0.97  & 14.65 &17.33 &26.70 & 27.32 & 25.22 \\
    \rowcolor{myhighlight}Keyboard & 0.00  & 7.46  & 0.00  & 1.07  & 2.21  & 0.46  & 46.01 & 18.12 & 51.17 &46.25 &90.25 & 84.05 & 80.65 \\
    \rowcolor{myhighlight}Display & 11.84 & 12.10 & 26.07 & 31.46 & 28.19 & 51.38 & 82.24 & 77.47 & 80.11 &78.20 &90.89 & 91.58 & 91.65 \\
    \rowcolor{myhighlight}Earphone & 2.59  & 6.62  & 0.92  & 4.52  & 22.85 & 20.82 & 3370.73 & 49.99 & 71.87 &61.81 &78.13 & 79.25 & 78.86 \\
    \rowcolor{myhighlight}Jar  & 2.10  & 13.77 & 2.96  & 2.29  & 21.22 & 22.42 & 33.81 & 21.27 & 22.18 &38.75 &39.66 & 33.20 & 33.15 \\
    \rowcolor{myhighlight}Knife & 0.00  & 0.00  & 0.00  & 0.00  & 1.00  & 0.00  & 0.00  & 0.00  & 1.40  &16.89 &27.70 & 26.66  & 26.57 \\
    \rowcolor{myhighlight}Lamp  & 13.97 & 27.06 & 25.67 & 43.75 & 52.96 & 62.17 & 89.10 & 73.67 & 80.54 &90.53 &89.47 & 90.68 & 90.48 \\
    \rowcolor{myhighlight}Laptop & 24.34 & 17.31 & 42.20 & 52.85 & 81.32 & 75.85 & 94.51 & 91.86 & 94.32 &97.04 &97.15 & 96.59 & 95.52 \\
    \rowcolor{myhighlight}Microphone & 0.00  & 0.00  & 0.00  & 0.01  & 0.01  & 0.06  & 0.00  & 0.02  & 0.96  &7.87 &1.38 & 2.26  & 2.45 \\
    \rowcolor{myhighlight}Microwave & 16.18 & 25.78 & 36.97 & 41.91 & 47.12 & 54.50 & 82.98 & 77.52 & 66.45 &87.76 &84.96 & 90.68 & 90.01 \\
    \rowcolor{myhighlight}Mug   & 2.60  & 7.84  & 4.74  & 5.00  & 13.03 & 27.83 & 83.00 & 42.79 & 73.64 &73.79 &92.40 & 93.15 & 92.73 \\
    \rowcolor{myhighlight}Printer  & 10.56 & 6.99  & 26.64 & 32.36 & 26.62 & 45.87 & 65.05 & 66.80 & 61.81 &66.61 &73.67 & 65.19 & 64.06 \\
    \rowcolor{myhighlight}Remote Control & 0.00  & 0.33  & 0.01  & 0.00  & 0.00  & 0.19  & 1.66  & 0.85  & 21.92 &12.52 &43.67 & 54.75 & 54.73 \\
    \rowcolor{myhighlight}Phone & 0.04  & 0.39  & 0.02  & 0.09  & 0.50  & 1.95  & 18.86 & 4.90  & 34.50 &21.68 &66.15 & 75.14 & 74.42 \\
    \rowcolor{myhighlight}Alarm & 0.49  & 4.37  & 0.85  & 2.44  & 6.84  & 7.79  & 35.16 & 17.02 & 26.70 &36.62 &43.81 & 49.04 & 50.49 \\
    \rowcolor{myhighlight}Book  & 3.91  & 5.70  & 6.59  & 5.72  & 8.87  & 19.42 & 54.52 & 30.53 & 57.04 &58.81 &70.69 & 75.29 & 75.12 \\
    \rowcolor{myhighlight}Cake  & 15.39 & 15.40 & 20.09 & 23.34 & 33.24 & 30.27 & 61.54 & 56.42 & 53.17 &64.27 &55.50 & 64.51 & 63.42 \\
    \rowcolor{myhighlight}Calculator & 0.02  & 0.19  & 0.68  & 0.05  & 0.61  & 2.56  & 16.88 & 10.28 & 27.70 &29.14 &46.45 & 51.23 & 50.27 \\
    \rowcolor{myhighlight}Candle & 11.18 & 10.59 & 11.51 & 18.20 & 17.60 & 36.10 & 53.74 & 36.69 & 26.06 &71.00 &69.85 & 69.29 & 69.32 \\
    \rowcolor{myhighlight}Charger & 0.01  & 0.01  & 0.07  & 0.00  & 0.01  & 0.71  & 10.35 & 0.90  & 12.03 &20.33 &38.89 & 49.32 & 46.48 \\
    \rowcolor{myhighlight}Chessboard & 0.00  & 3.19  & 0.11  & 8.91  & 36.56 & 45.26 & 78.72 & 62.71 & 67.45 &61.24 &70.46 & 76.69 & 78.07 \\
    \rowcolor{myhighlight}Coffee\_Machine & 11.13 & 20.39 & 20.38 & 18.43 & 34.55 & 53.51 & 77.56 & 57.93 & 38.69 &56.74 &49.34 & 60.68 & 60.85 \\
    \rowcolor{myhighlight}Comb  & 0.00  & 0.00  & 0.00  & 0.04  & 0.08  & 0.01  & 5.77  & 0.03  & 2.13  &19.59 &35.99 & 43.72 & 37.76 \\
    \rowcolor{myhighlight}Cutting\_Board & 2.79  & 0.69  & 2.19  & 4.57  & 19.23 & 15.97 & 0.00  & 27.23 & 54.40 &61.25 &49.67 & 63.59 & 55.86 \\
    \rowcolor{myhighlight}Dishes & 1.30  & 2.28  & 1.73  & 5.15  & 10.01 & 15.77 & 65.85 & 29.00 & 61.69 &77.31 &72.25 & 71.16 & 68.38 \\
    \rowcolor{myhighlight}Doll  & 0.14  & 1.01  & 0.00  & 6.84  & 5.52  & 2.06  & 9.12  & 1.44  & 14.48 &15.49 &10.18 & 28.82 & 25.03 \\
    \rowcolor{myhighlight}Eraser & 0.00  & 0.00  & 0.00  & 0.00  & 0.00  & 0.00  & 0.00  & 0.00  & 10.29 &3.41 &19.20 & 35.99 & 36.67 \\
    \rowcolor{myhighlight}Eye\_Glasses & 0.15  & 1.78  & 0.65  & 0.48  & 2.96  & 9.69  & 72.73 & 37.83 & 77.15 &71.12 &90.49 & 95.52 & 95.21 \\
    \rowcolor{myhighlight}File\_Box & 15.22 & 15.48 & 23.51 & 26.95 & 35.66 & 53.42 & 63.36 & 62.91 & 66.86 &70.03 &73.19 & 71.31 & 71.46 \\
    \rowcolor{myhighlight}Fork  & 0.00  & 0.00  & 0.00  & 0.00  & 0.17  & 0.00  & 0.05  & 0.00  & 2.98  &5.59 &8.86 & 14.91 & 14.34 \\
    \rowcolor{myhighlight}Fruit & 0.77  & 0.88  & 0.41  & 1.82  & 7.11  & 11.44 & 58.14 & 22.36 & 42.00 &59.38 &76.82 & 77.79 & 78.10 \\
    \rowcolor{myhighlight}Globe & 25.48 & 25.42 & 16.80 & 37.32 & 49.97 & 32.95 & 75.62 & 64.77 & 64.98 &74.12 &80.54 & 79.33 & 78.30 \\
    \rowcolor{myhighlight}Hat   & 0.04  & 13.90 & 2.46  & 1.66  & 3.41  & 6.23  & 46.89 & 24.34 & 32.08 &40.68 &50.14 & 52.34 & 49.58 \\
    \rowcolor{myhighlight}Mirror & 0.13  & 16.71 & 0.03  & 0.00  & 0.00  & 0.90  & 0.47  & 3.76  & 0.97  &23.89 &1.17 & 2.61  & 2.30 \\
    \rowcolor{myhighlight}Notebook & 0.31  & 0.95  & 1.19  & 0.41  & 2.04  & 6.87  & 24.77 & 7.09  & 45.73 &25.23 &52.65 & 64.35 & 63.36 \\
    \rowcolor{myhighlight}Pencil & 0.00  & 0.03  & 0.00  & 0.00  & 0.02  & 0.03  & 0.00  & 0.21  & 7.83  &13.81 &23.23 & 39.72 & 40.72 \\
    \rowcolor{myhighlight}Plant & 12.55 & 25.19 & 24.01 & 39.13 & 49.13 & 52.49 & 84.15 & 66.91 & 75.74 &83.58 &86.05 & 86.11 & 85.76 \\
    \rowcolor{myhighlight}Plate & 0.36  & 0.69  & 0.41  & 0.87  & 1.61  & 16.54 & 38.90 & 18.06 & 74.66 &58.51 &94.49 & 96.05 & 95.99 \\
    \rowcolor{myhighlight}Radio & 0.07  & 0.13  & 2.14  & 0.02  & 1.85  & 3.39  & 4.81  & 3.18  & 11.42 &10.58 &16.00 & 16.58  & 13.37 \\
    \rowcolor{myhighlight}Ruler & 0.00  & 0.06  & 0.00  & 0.00  & 0.00  & 0.00  & 0.17  & 0.07  & 8.07  &6.75 &23.11 & 38.20 & 37.26 \\
    \rowcolor{myhighlight}Saucepan & 21.64 & 16.14 & 11.19 & 26.24 & 17.81 & 31.15 & 73.49 & 30.25 & 24.75 &48.26 &50.70 & 63.28 & 63.36 \\
    \rowcolor{myhighlight}Spoon & 0.00  & 0.00  & 0.00  & 0.00  & 0.10  & 0.20  & 0.56  & 1.21  & 5.45  &9.08 &17.86 & 22.30 & 20.90 \\
    \rowcolor{myhighlight}Tea\_Pot & 7.03  & 12.42 & 9.29  & 21.50 & 31.24 & 34.85 & 87.60 & 73.44 & 76.46 &88.12 &88.04 & 91.11 & 91.24 \\
    \rowcolor{myhighlight}Toaster & 2.18  & 6.18  & 10.51 & 5.12  & 4.86  & 15.79 & 32.65 & 26.29 & 27.72 &35.77 &24.27 & 34.40 & 34.52 \\
    \rowcolor{myhighlight}Vase & 7.43  & 17.93 & 11.02 & 19.63 & 36.39 & 33.34 & 64.57 & 52.35 & 52.75 &63.92 &67.60 & 73.40 & 73.03 \\
    \rowcolor{myhighlight}Vegetables & 0.00  & 0.18  & 0.13  & 0.20  & 4.91  & 9.75  & 0.01  & 5.00  & 9.62  &9.34 &7.05 & 14.00  & 13.44 \\

      \bottomrule
    \end{tabular}
  \label{tbl:toscene-percls-map50}
\end{table*}

% ---- Bibliography ----
%
% BibTeX users should specify bibliography style 'splncs04'.
% References will then be sorted and formatted in the correct style.
%
\bibliographystyle{splncs04}
\bibliography{main}
\end{document}